\documentclass[journal,comsoc]{IEEEtran}
\usepackage{float}
\usepackage[T1]{fontenc}
\ifCLASSINFOpdf
\else
\fi
\usepackage{hyperref}
\usepackage{amsmath}
\usepackage{color}
\usepackage[dvipsnames]{xcolor}
\usepackage[export]{adjustbox}
\usepackage{color}
\usepackage{graphicx}
\usepackage{booktabs}
\usepackage{multirow}
\usepackage{array}
\usepackage{caption}
\usepackage{amssymb}
\usepackage{pifont}
\usepackage{comment}
\usepackage{makecell}

\usepackage{enumitem}
\usepackage{cite}
\usepackage{makecell}
\usepackage{url}
\usepackage{tikz}
\usetikzlibrary{positioning, shapes, arrows.meta}
\newcolumntype{L}{>{\centering\arraybackslash}m{0.8cm}}
\makeatletter
\newcommand\footnoteref[1]{\protected@xdef\@thefnmark{\ref{#1}}\@footnotemark}
\makeatother
\usepackage{footnote}
\makesavenoteenv{tabular}
\makesavenoteenv{table}

\usepackage{algorithm}
\usepackage{algorithmic}

\begin{document}

\title{Clear Roads, Clear Vision: Advancements in Multi-Weather Restoration for Smart Transportation}

\author{Vijay M. Galshetwar, Praful Hambarde, Prashant W. Patil, Akshay Dudhane, and Sachin Chaudhary
\thanks{Vijay M. Galshetwar is with the Finolex Academy of Management and Technology, Ratnagiri, Maharashtra, India (e-mail: vijay.galshetwar@famt.ac.in). \newline Praful Hambarde is with Drone Lab, Centre for Artificial Intelligence and Robotics, IIT Mandi, India (e-mail: praful@iitmandi.ac.in). \newline Prashant W. Patil is with Mehta Family School of Data Science and Artificial Intelligence, IIT Guwahati, India (e-mail: pwpatil@iitg.ac.in). \newline Akshay Dudhane is with the SPACE42, Abu Dhabi, United Arab Emirates (e-mail: akshay.dudhane@space42.ai). 
\newline Sachin Chaudhary is with the Centre of Excellence: Artificial Intelligence, School of Computer Science, UPES Dehradun, India. (E-mail: sachin.chaudhary@ddn.upes.ac.in).
\newline This paper has been accepted for publication in IEEE Transactions on Intelligent Transportation Systems (IEEE). © IEEE. Personal use is permitted. Permission from IEEE must be obtained for all other uses.}}

\markboth{\textit{IEEE Transactions on Intelligent Transportation Systems}}%
{Shell \MakeLowercase{\textit{et al.}}: Bare Demo of IEEEtran.cls for IEEE Communications Society Journals}
\maketitle

\begin{abstract}
    Adverse weather conditions such as haze, rain, and snow significantly degrade the quality of images and videos, posing serious challenges to intelligent transportation systems that rely on visual input. These degradations affect critical applications including autonomous driving, traffic monitoring, and surveillance. This survey presents a comprehensive review of image and video restoration techniques developed to mitigate weather-induced visual impairments. We categorize existing approaches into traditional prior-based methods and modern data-driven models, including CNNs, transformers, diffusion models, and emerging vision-language models. Restoration strategies are further classified based on their scope: single-task models, multi-task/multi-weather systems, and all-in-one frameworks. In addition, we discuss day and night time restoration challenges, benchmark datasets, and evaluation protocols. 
    The survey concludes by discussing current limitations and future directions, including unified restoration with downstream perception, real-time video restoration, and benchmarks for compound degradations under dynamic lighting. This work aims to serve as a valuable reference for advancing weather-resilient vision systems in smart transportation environments. Lastly, to keep pace with the rapid progress in this area, we will regularly update the latest relevant papers and their open-source implementations \href{https://github.com/ChaudharyUPES/A-comprehensive-review-on-Multi-weather-restoration}{here.}
\end{abstract}

\vspace{4mm}
\section{Introduction}
\begin{figure}[t]
\centering
  \includegraphics[width=0.95\linewidth]{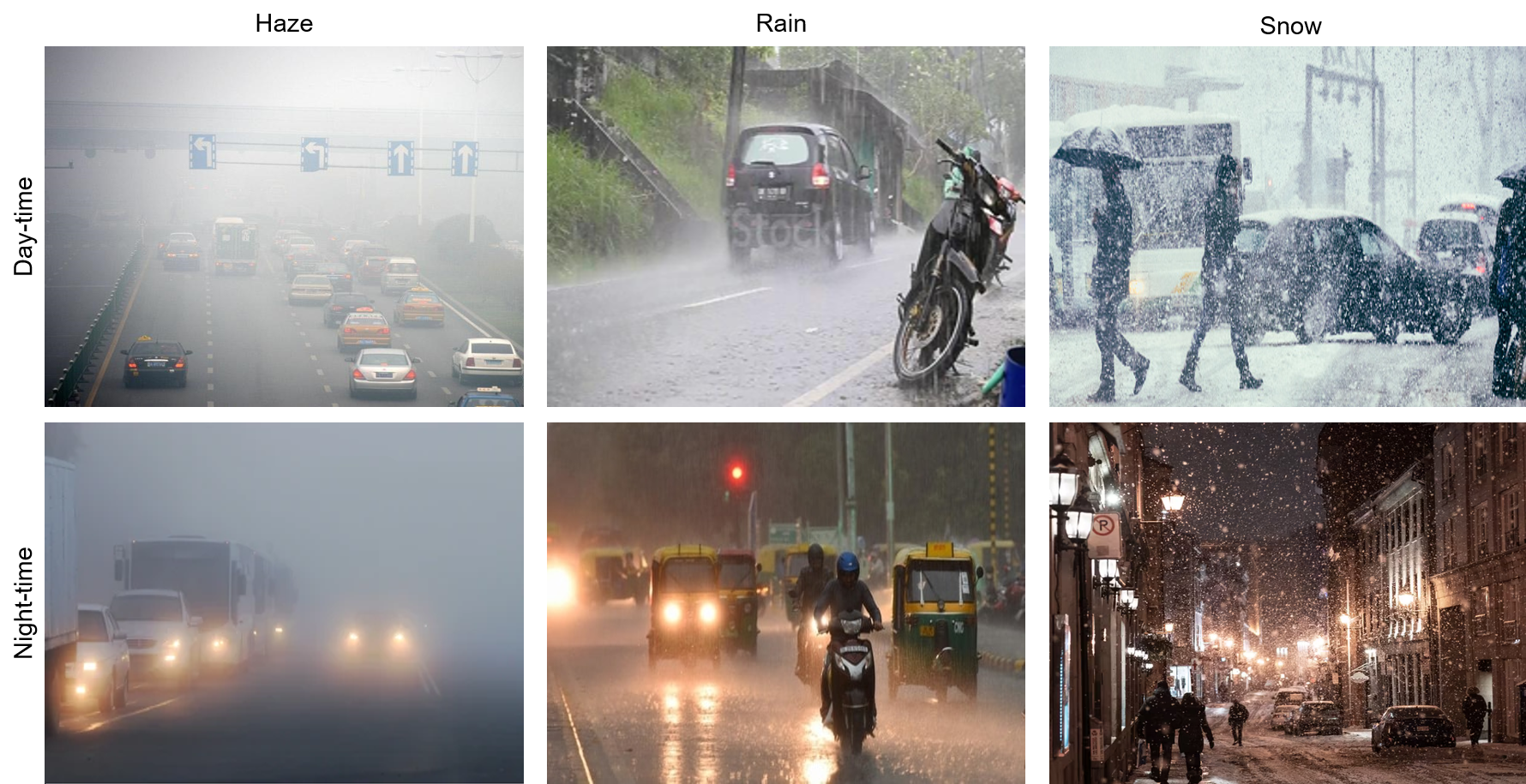}
   \caption{Sample real-world degradations in images.}
\label{FIG:real}
\end{figure}

\IEEEPARstart{I}{n} modern intelligent transportation systems (ITS), computer vision plays a pivotal role in enabling tasks such as lane detection, object tracking, autonomous driving, traffic monitoring, and autonomous navigation. These systems depend on clear and reliable visual data from on-board cameras and roadside infrastructure. However, adverse weather such as haze, rain, snow, and fog degrades visibility and scene understanding, reducing downstream perception accuracy and increasing risk in safety-critical settings~\cite{kulkarni2022unified, zhu2024multi, gautam2025pureformer}. Such conditions introduce visibility loss, occlusion, noise, and distortions that obscure key details and hinder vision algorithms~\cite{kulkarni2022unified}. Haze reduces contrast and color fidelity, rain adds streaks and blur, and snow causes occlusions and texture-like noise. Figure \ref{FIG:real} illustrates the real-world weather degraded images~\cite{dai2023mipi, alaspure2020darkgan}. Each weather type introduces distinct distortions, making generalization challenging.
%
As a result, multi-weather image/video restoration has become a critical pre-processing step for robust transportation systems, aiming to recover clear images and videos degraded by haze, rain, snow, and other atmospheric conditions~\cite{kang2018deep, siddiqua2021multi, patil2022video, kulkarni2022unified, kulkarni2022wipernet, dudhane2020varicolored, musat2021multi, zhu2024multi, ai2024multimodal}. Traditional methods rely on scattering models and handcrafted priors~\cite{wang2017hierarchical, zandersen2021nature}, but often fail in complex real-world settings. Recent deep learning approaches (CNNs~\cite{zhang2020multi, fan2023semi, kulkarni2022unified}, GANs~\cite{wan2019perceptual, qian2018attentive}, transformers~\cite{valanarasu2022transweather}, distillation~\cite{9879902}, domain translation~\cite{patil2023multi}, prompt learning~\cite{ai2024multimodal}, and diffusion models~\cite{ozdenizci2023restoring}) better capture weather-specific patterns and improve visibility across diverse conditions. Key applications and challenges are as below: 

\subsection{Significance of Multi-weather Restoration in ITS}
\begin{itemize}
    \item \textbf{Enhancing driver assistance systems:} In autonomous driving, restoration techniques enhance visibility in bad weather, ensuring reliable visual data. \cite{siddiqua2021multi, patil2022video, kumar2020efficient}.
    \item \textbf{Improving transportation monitoring:} In traffic monitoring centers, under severe weather, video restoration is crucial for reliable and secure operation \cite{kumar2020efficient, musat2021multi}.
    \item \textbf{Airport and port operations:} Image restoration improves runway visibility for air traffic control and ensures clear visuals for safe vessel docking and navigation \cite{musat2021multi}. 
\end{itemize}
\subsection{Major Challenges of Multi-weather Restoration:}
\begin{figure}[t]
    \centering
      \includegraphics[width=1\linewidth]{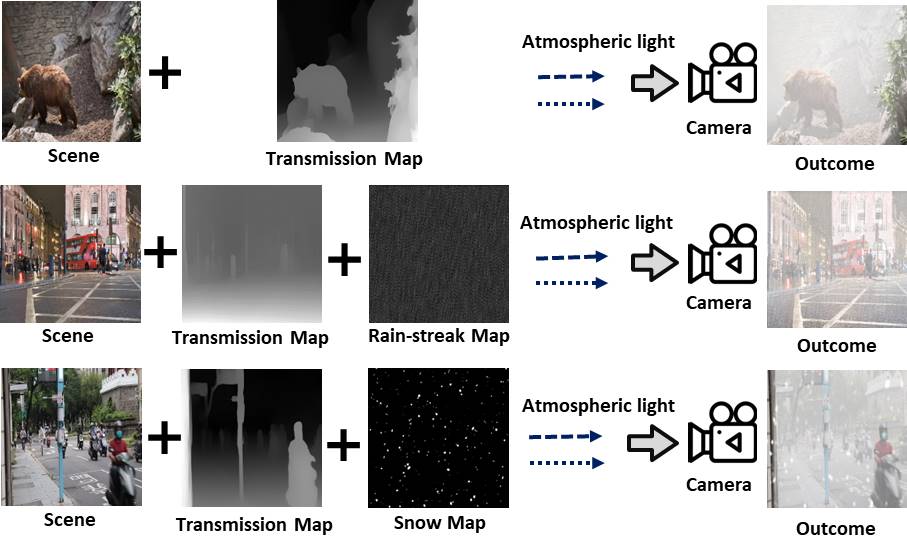}
       \caption{Synthetic pipeline: atmospheric light is combined with (i) a transmission map for depth-dependent haze, (ii) rain-streak overlays, and (iii) snow-particle maps to generate realistic hazy, rainy, and snowy scenes.}
    \vspace{-5mm}
    \label{FIG:Oview}
    \end{figure}

    A major challenge in multi-weather restoration is the lack of real-world datasets due to safety, cost, and environmental constraints. As a result, most works rely on synthetic data to simulate weather, as illustrated in Figure~\ref{FIG:Oview} for haze, rain, and snow. Other key challenges include:
    \begin{itemize}
        \item \textbf{Complex and mixed weather conditions:} Real-world scenes often involve combinations of fog, rain, and snow, making restoration more complex~\cite{10571568, dudhane2024dynamic, dudhane2022burst}.
        \item \textbf{Task interference in multi-task learning:} Jointly learning multiple restoration tasks often introduces performance trade-offs~\cite{kang2018deep}.
        \item \textbf{Computational efficiency:} Real-time requirements for ITS demand lightweight and efficient models~\cite{kulkarni2022unified}.
        \item \textbf{Lack of diverse and high-quality datasets:} Synthetic datasets often lack real-world diversity~\cite{ai2024multimodal}.
        \item \textbf{Underexplored multi-weather video restoration:} Despite advances in image restoration, video-based methods remain largely unexplored, except few attempts~\cite{patil2022video, patil2022robust}.
    \end{itemize}

    \noindent This survey reviews state-of-the-art (SOTA) multi-weather image and video restoration methods for dehazing, deraining, and desnowing, covering prior-based and learning-based approaches, including unified models. We also discuss their importance for smart transportation, where reliable visibility is critical for decision-making. \textbf{Our key contributions are:}
    \begin{itemize}
        \item Provide a comprehensive overview of methods for image/video dehazing, deraining, desnowing, and unified multi-weather restoration.
        \item Summarize SOTA approaches, including prior-based models, learning-based methods, and hybrid techniques.
        \item Discussed evaluation metrics and benchmark datasets relevant to image/video restoration in ITS.
        \item Identify current challenges and open research directions for advancing multi-weather restoration algorithms.
    \end{itemize}

    \noindent Through this overview, we aim to support researchers in the development of robust, weather-resilient vision systems for next-generation smart transportation.

\section{Related Work: Weather-specific Approaches}
\label{sec:related work}
\begin{figure*}[t]
\centering
  \rotatebox{-90}{\includegraphics[width=0.63\linewidth]{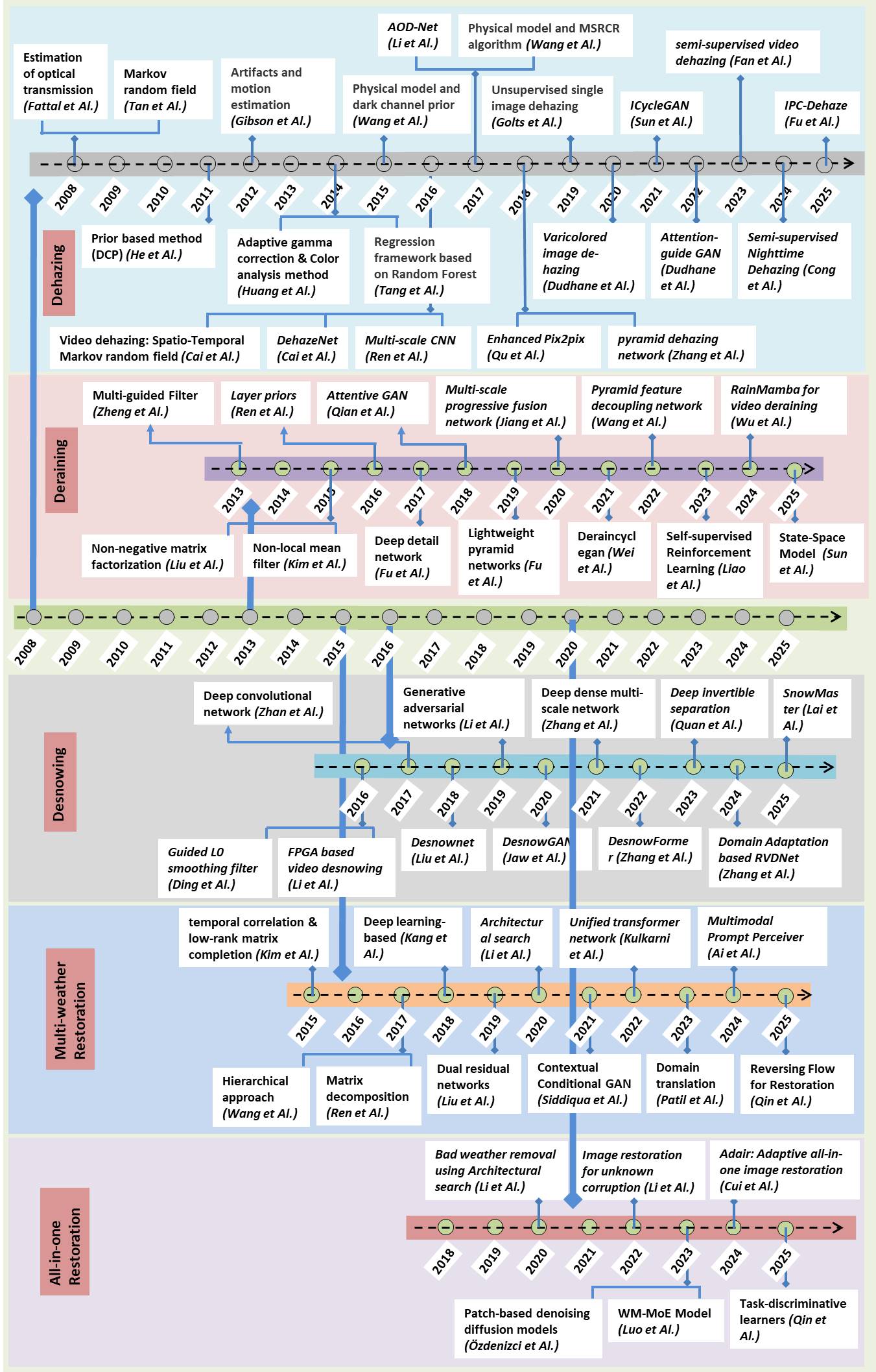}}
   \caption{The development timeline of hazy, rainy, snowy and multi-weather degraded image/video restoration approaches.}
\vspace{-1mm}
\label{FIG:Timeline}
\end{figure*}

    The literature on restoration techniques is categorized into five main areas: haze, rain, snow, multi‑weather restoration, and all‑in‑one, approaches. We additionally discussed existing traffic-sign interpretation approaches, as this application supports decision-making in autonomous systems for intelligent transportation systems. This section gives an overview of the first three categories \textit{i.e. weather specific} and traffic sign interpretation, including traditional and learning-based approaches. The detailed review of existing multi-weather restoration and all-in-one approaches is provided in Sections~\ref{sec: multi-weather restoration} and ~\ref{sec: all-in-one}, respectively. 
    Figure \ref{FIG:Timeline} illustrates a timeline of major image/video restoration methods.

\subsection{De-hazing Approaches}
    \vspace{-0.55mm}
    The formation of synthetic hazy image using atmospheric scattering model \cite{5567108} as: $I_{x}(n) = J_{x}(n)\cdot T_{x}(n) + A\cdot(1-T_{x}(n))$
    \noindent 
    where, $\displaystyle {I_{x}(n)}$ and $\displaystyle {J_{x}(n)}$ are the hazy and haze-free images at pixel $\displaystyle {x}$ and time $\displaystyle {n}$ respectively, $\displaystyle {A}$ is the atmospheric light, and $\displaystyle {T_{x}(n)}$ is scene transmission map of the image estimated as $\displaystyle {T_{x}(n) = e^{(-\beta d(x))}}$, where, $\displaystyle {\beta}$ denotes attenuation coefficient and $\displaystyle {d(x)}$ denotes depth of the scene.

\subsubsection{\textbf{Image De-hazing}}
    Major categories are as below:\newline
    \textit{\underline{Prior-based Methods:}} Wang~\textit{et al.}~\cite{wang2015single} proposed a physical-model-based de-hazing technique using the dark channel prior (DCP) with atmospheric light estimation. They later enhanced it with a multi-scale retinex and color restoration scheme~\cite{wang2017single}. Other notable methods include linear transformation~\cite{berman2018single}, detail manipulation~\cite{li2015single}, and confidence priors~\cite{yuan2021confidence} to improve performance in complex scenes. A unified model for bridging haze scenarios was also introduced in~\cite{feng2024bridging}, offering improved generalization and stability.\\
    \underline{\textit{Filter based Methods:}} 
    A multi-scale correlated wavelet approach is proposed by Liu \textit{et al.}~\cite{liu2017efficient} for simultaneous de-hazing and denoising. A globally guided image filtering technique was introduced by Li \textit{et al.}~\cite{li2017_single} for contrast enhancement and high-quality restoration. Ma \textit{et al.}~\cite{ma2023image}, improved color channel transfer and multiexposure fusion with k-means clustering were employed for effective de-hazing.\\    
    \underline{\textit{Markov Random Field Based Methods:}} Tan \textit{et al.} \cite{tan2008visibility} developed a cost function in the framework of Markov random fields which can be efficiently optimized by various techniques, such as graph-cuts or belief propagation.\\
    \underline{\textit{Learning-based Methods:}} Galdran~\textit{et al.}\cite{galdran2015enhanced} proposed a variational de-hazing framework, while Cai \textit{et al.}\cite{cai2016dehazenet} introduced a CNN-based end-to-end method. Liu \textit{et al.}\cite{liu2019griddehazenet} designed an attention-driven multi-scale network for fast and accurate de-hazing. Dudhane \textit{et al.}~\cite{dudhane2020varicolored} developed a varicolored network to restore color balance in hazy images. 
    Zhang \textit{et al.}\cite{zhang2020pyramid} proposed several strategies, including a pyramid channel-based framework and a multi-level feature enhancement method \cite{zhang2020multi}. Zhu \textit{et al.} \cite{zhu2020novel} introduced a multi-exposure fusion technique, while Shyam \textit{et al.} \cite{shyam2021towards} focused on domain-invariant de-hazing. Bai \textit{et al.} \cite{bai2022self} proposed a progressive feature refinement strategy. Additionally, Dudhane \textit{et al.}\cite{8802288, 9025402} proposed deep fusion and residual inception-based GAN models for improved dehazing, while Yu \textit{et al.} \cite{yu2024vifnet} introduced visible–infrared fusion for enhanced visibility.

    Beyond paired data settings, Engin \textit{et al.} \cite{engin2018cycle} adopted an unpaired training scheme using CycleGAN for flexible de-hazing. Zhu \textit{et al.} \cite{zhu2018dehazegan} incorporated the atmospheric scattering model into a GAN framework to improve visual quality. Dudhane \textit{et al.} \cite{dudhane2019cdnet} introduced single image de-hazing using unpaired adversarial training.  Ren \textit{et al.} \cite{ren2020single} further refined the process by incorporating holistic edge information with multi-scale CNN. Wang \textit{et al.} \cite{wang2021tms} proposed TMS-GAN to mitigate domain shifts between synthetic and real-world hazy images. Wang \textit{et al.} \cite{9766195} developed a cycle spectral normalized soft likelihood estimation patch GAN for haze removal, while Manu \textit{et al.} \cite{manu2023ganid} presented GANID for high-contrast, color-preserving de-hazing across natural and synthetic datasets. Song \textit{et al.} \cite{song2023vision} proposed DehazeFormer that consists of modified normalization layer, activation function, and spatial information aggregation scheme. Further, Liu \textit{et al.} \cite{liu2024self} developed a self-enhancement GAN algorithm incorporating depth estimation. Li \textit{et al.}\cite{li2022image} approached de-hazing as a two-way image translation problem using a weakly supervised framework. Along with homogeneous de-hazing, authors introduced with a Self-paced Semi-Curricular attention Network by Guo \textit{et al.} \cite{guo2023scanet} and image processing network by Kim \textit{et al.} \cite{kim2023deep} for non-homogeneous de-hazing. All these above methods are purely trained with synthetically generated data, which limits the performance on real-world hazy data. Wei \textit{et al.} \cite{wei2024robust} presented a robust unpaired image de-hazing approach with adversarial deformation constraints to align hazy and clean image distributions. Fu \textit{et al.} \cite{Fu_2025_CVPR} proposed IPC-Dehaze is an iterative predictor-critic code decoding for real-world image de-hazing. Liu \textit{et al.} \cite{10946240} presented a novel variational nighttime de-hazing framework using hybrid regularization that enhances the perceptual visibility of nighttime hazy scene.\\
    Li \textit{et al.} \cite{li2019semi} proposed a semi-supervised learning network for image de-hazing that combines synthetic and real-world hazy images, enhancing the model's generalization through supervised and unsupervised techniques. Cong \textit{et al.} \cite{cong2024semi} introduced a semi-supervised nighttime de-hazing method with spatial-frequency awareness and realistic brightness constraints.\\
    Wu \textit{et al.} \cite{wu2021contrastive} developed a compact single-image de-hazing network utilizing contrastive learning. Yang \textit{et al.} \cite{yang2022self} introduced a self-augmented unpaired de-hazing method that uses density and depth decomposition, addressing limitations in synthetic paired training data requirement. Ding \textit{et al.} \cite{ding2023u2d2net} developed a unified de-hazing and denoising network using DCP with an edge-aware network. Wei \textit{et al.} \cite{wei2024robust} introduced an adversarial deformation constraint for robust unpaired image de-hazing. Wang \textit{et al.} \cite{wang2024ucl} developed an unsupervised contrastive learning framework that trains on unpaired clean and hazy images. Wang \textit{et al.} \cite{WANG2024109956} proposed a contrastive learning based hazy weather restoration network.

\subsubsection{\textbf{Video De-hazing}}
    Major categories are as below: \newline
    \textit{\underline{Prior-Based Methods:}} Park \textit{et al.} \cite{park2017video} introduced a video de-hazing system leveraging fast airlight estimation and the DCP to maintain temporal coherence across frames, improving video quality and visibility. Dong \textit{et al.} \cite{dong2019efficient} developed an adaptive DCP method that incorporates spatial-temporal correlations for real-time traffic video de-hazing. Adidela \textit{et al.} \cite{adidela2021single} consolidated state-of-the-art DCP-based techniques for both single-image and video de-hazing. Wu \textit{et al.} \cite{wu2024real} proposed a real-time HD video defogging approach using a modified DCP algorithm, tailored for high-definition content and suitable for immediate application. The artifacts issue is tackled in Li \textit{et al.} \cite{li2020zero} with gradient and color prior regularization and Ashwini \textit{et al.} \cite{ashwini2022image} with improved gradient preservation. Some contrast enhancement methods like segments videos into view-based clusters Yu \textit{et al.} \cite{yu2016image}, transmission estimation using the HSL color model Soma \textit{et al.} \cite{soma2022efficient}, physical priors and temporal information across video frames Xu \textit{et al.} \cite{xu2023video} and dual-transmission-map Auoub  \textit{et al.} \cite{ayoub2024video} are proposed.
    \newline
    \textit{\underline{Markov Random Field based Methods:}} Zhang \textit{et al.} \cite{zhang2011video} and Cai \textit{et al.} \cite{cai2016real} presented the haze-free video model by assuming frame-by-frame video sequences for improving temporal coherence utilizing Markov Random Field and optical flow.
    \newline
    \textit{\underline{Retinex Theory Based Methods:}} Xue \textit{et al.} \cite{xue2016video} introduced a video de-hazing algorithm that utilizes Multi-Scale Retinex with Color Restoration.
    \newline
    \textit{\underline{Learning-based Methods:}} Fan xue \cite{fan2023semi} introduced a semi-supervised video de-hazing method leveraging CNNs and a dynamic haze generator. Galshetwar \textit{et al.} proposed various approaches for video de-hazing primarily focusing on computational complexity and temporal consistency aspect \cite{galshetwar2021single, galshetwar2023lrnet, galshetwar2024drfnet, galshetwar2024cross}.
    \newline
    \textit{\underline{Non-homogeneous Methods:}} Ancuti \textit{et al.} \cite{ancuti2023ntire} reported on the NTIRE 2023 HR nonhomogeneous de-hazing challenge, showcasing advancements and benchmarking state-of-the-art methods for high-resolution videos with complex haze distributions. Liu \textit{et al.} \cite{liu2023ntire} emphasized the quality assessment of video enhancement techniques in the same challenge, offering a detailed evaluation framework for nonhomogeneous de-hazing and encouraging the development of more robust de-hazing algorithms.

\subsection{De-raining Approaches}
    \vspace{-1mm}
    For supervised training, getting rainy and rain-free images is difficult. Therefore, researchers started training the supervised models with synthetic data provided in Yang \textit{et al.} \cite{9439949}. Mathematically, the rainy data is generated as: 
    \begin{equation}
        O_t =  B_t + S_t,\,\,\,\,\,\,\,\,\,\,\,\,\,\   t = 1,2,...,N
        \label{Eq:rain_video1}
    \end{equation}
    where, $S_t$ represents rain streaks of $t^{th}$ frame, $B_t$ represent the $t^{th}$ rain-free frame, $O_t$ is the $t^{th}$ synthetically generated rainy frame and $t$ is the temporal indicator, $N$ denotes the number of video frames. With the introduction of rain accumulation and accumulation flow the above equation is expressed as:
    \begin{equation}
     O_t =  T_t B_t + (1-T_t) A_t + U_t + S_t, \,\,\,\,\,\,\,\,\,\,\,\,\,\  t = 1,2,...,N.
    \label{Eq:rain_video2}
    \end{equation}
    where, $A_t$ is the global atmospheric light, $T_t$ is atmospheric transmission map, $U_t$ is rain accumulation flow layer based on atmospheric flow and local raindrop density.
    \begin{equation}
     O'_t =  (1-\alpha_t) (B_t + S_t) +  \alpha_t R_t
    \label{Eq:rain_video3}
    \end{equation}
    where, $\alpha_t$ is an alpha matting map, and $R_t$ is the rain reliance map. Using Eq. \ref{Eq:rain_video1}-\ref{Eq:rain_video3}, rain model with occlusion is given as:
    \begin{equation}
     O'_t =  (1-\alpha_t) (O_t) + \alpha_t R_t
    \label{Eq:rain_video4}
    \end{equation}
    Hence, Eq. \ref{Eq:rain_video4} represents a rain model that captures rain streaks, accumulation, accumulation flow, and occlusions in a comprehensive way \cite{9439949}.

\subsubsection{\textbf{Image De-raining}}
    Major categories are as below:
    \noindent \textit{\underline{Prior-based Methods:}} 
    Luo \textit{et al.} \cite{luo2015removing} introduced a discriminative sparse coding approach that utilizes dictionary learning to separate rain streaks from the background, preserving image details effectively. Li \textit{et al.} \cite{li2016rain} proposed a layer-based model using priors to guide the separation of rain streaks as a distinct layer. Focusing on gradient domain analysis, \cite{du2018single} presented a method that decorrelates rain streaks and background by leveraging the differential impact of rain streaks on the X- and Y-gradients of an image, producing visually clear outputs.
    \newline
    \textit{\underline{Filter Based Methods:}} \cite{kim2015multi} proposed a multi-frame de-raining algorithm that employs a motion-compensated non-local mean filter to enhance rain removal in dynamic video scenes. \cite{santhaseelan2015utilizing} introduced a  guided filtering to preserve background textures and edges, making it suitable for real-time use. For static images, \cite{gautam2018rain} developed a guided filter-based approach to reduce noise from rain and snow particles.
    \newline
    \textit{\underline{Matrix Decomposition Based Methods:}}  \cite{ren2017video} introduced a matrix decomposition-based method for video de-snowing and de-raining, which can also be applied to single images. \cite{liu2015rain} proposed a rain removal method using non-negative matrix factorization (NMF) for single images.
    \newline
    \textit{\underline{Learning-based Methods:}}
    Li \textit{et al.} \cite{li2018recurrent} introduced a recurrent squeeze-and-excitation context aggregation network for image de-raining. A context aggregation based network is proposed in \cite{deng2020detail}. Jiang \textit{et al.} \cite{jiang2020multi} developed a multi-scale progressive fusion network to refine images at multiple scales for rain streak removal. Yang \textit{et al.} \cite{yang2017deep} proposed a deep joint rain detection and removal framework that employs a CNN to detect and eliminate rain streaks simultaneously. A adversarial learning-based approaches are proposed \cite{zhang2019image}, \cite{wei2021deraincyclegan} for image de-raining. Fu \textit{et al.} \cite{fu2019lightweight} introduced lightweight Laplacian pyramid decomposition network for image de-raining to achieve high-quality results with low computational complexity. Further, the pyramid feature decoupling network is proposed in \cite{wang2022pfdn}, which enhances image clarity by decoupling multi-scale features. Xiao \textit{et al.} \cite{xiao2022image} proposed the Image De-raining Transformer, that incorporates general priors of vision tasks, such as locality and hierarchy, into the network design. Li \textit{et al.} \cite{10359459} presented image de-raining via similarity diversity model for single traffic. For lane detection and depth estimation, Li \textit{et al.} \cite{10786924} proposed a ultra-fast de-raining plugin for vision-based perception of autonomous driving.

\subsubsection{\textbf{Video De-raining}}
     \textit{\underline{Prior-based Methods:}} \cite{kim2015multi} proposed a multi-frame de-raining algorithm using a motion-compensated non-local mean filter for rainy video sequences. \cite{santhaseelan2015utilizing} introduced a method to Utilize local phase information to remove rain from video. Islam \textit{et al.} \cite{9656178} proposed a video de-raining considering the visual properties of rain streaks.
\newline
\textit{\underline{Learning-based Methods:}} Mi \textit{et al.} \cite{mi2016image} developed an image fusion-based video de-raining method using sparse representation. A progressive subtractive recurrent lightweight network is proposed in \cite{9645369}. Further, multi-patch progressive neural network is presented in \cite{10122854}. Semi-supervised approach with dynamical rain generator is proposed in \cite{9577812}. Yang \textit{et al.} \cite{yang2019frame} proposed a two-stage recurrent network with dual-level flow regularizations to perform the inverse recovery process of the rain synthesis model. Yan \textit{et al.} \cite{yan2021self} proposed a self-alignment network with transmission-depth consistency. Zhang \textit{et al.} \cite{9706330} introduced enhanced spatio-temporal interaction learning for Video Deraining. Wang \textit{et al.} \cite{wang2023unsupervised} presented a novel approach by integrating a bio-inspired event camera into the unsupervised video de-raining pipeline, which captures high temporal resolution information and model complex rain characteristics. A hybrid transformer with global and local representations is developed in \cite{mao2024aggregating}. Lin \textit{et al.} \cite{lin2024nightrain} 
introduced nighttime video de-raining method with adaptive rain removal and adaptive correction. Wu \textit{et al.} \cite{wu2024rainmamba} proposed an improved state space models based video de-raining network (RainMamba) with a novel Hilbert scanning mechanism to capture sequence level local information. Semi-supervised state-space model with dynamic stacking filter is proposed by Sun \textit{et al.} \cite{Sun_2025_CVPR} for real-world video de-raining.
\subsection{De-snowing Approaches}
The de-snowing model training and evaluation is based on synthetic dataset as getting paired clean and snowy data is difficult \cite{Chen_2023_ICCV}. The synthetic data is generated as:
\begin{equation}
 \alpha (x) = \left\{\begin{matrix}
\sigma \left ( -\left ( V(x)-\gamma  \right ) \times \beta \right ), \,\,\,\,\,\,\,\,\ if \, \, daytime \\
\sigma \left ( \left ( V(x)-\gamma  \right ) \times \beta \right ), \,\,\,\,\,\,\, if \, \, nighttime
\end{matrix}\right.
\label{Eq:snow_video1}
\end{equation}
\cite{Chen_2023_ICCV} first convert the RGB image to the HSV color space, where V = max(R, G, B) represents the
largest color component. v denotes the value of the V channel in the HSV space, which is normalized to the range of 0 to 1. $\gamma$ and $\beta$ are adjusted based on the specific video, and $\sigma$ represents the softmax function. The formation of snowy video is as below:
$Z(x) = I_{haze}(x) + \alpha (x) Aug(S(x))$
where \textit{Aug} denotes data augmentation. $Z$ is the final output with both snow and haze.

\subsubsection{\textbf{Image De-snowing}}
    Major categories are as below.
    
    \noindent \textit{\underline{Prior-based Methods:}} Zhang \textit{et al.} \cite{zhang2021deep} proposed a deep dense multi-scale network for snow removal, utilizing semantic and depth priors to enhance image quality. A hierarchical dual-tree complex wavelet representation and contradict channel loss is proposed in \cite{chen2021all} to improve the performance.
    \newline
    \textit{\underline{Filter Based Methods:}} A guided smoothing filter was proposed in~\cite{ding2016single} for single image rain and snow removal. In~\cite{farhadifard2017single}, a supervised median filtering scheme was introduced for marine snow removal. A snowfall model smoothing filter that preserves edge features was presented in~\cite{fan2018preservation}.
    \newline
    \textit{\underline{FPGA Based Methods:}} \cite{li2016snow} presented an FPGA-based snow removal approach capable of real-time processing for images with a minimum resolution of 640$\times$480, demonstrating the practicality of hardware-accelerated de-snowing solutions.
    \newline
    \underline{\textit{Learning-based Methods:}} Zhan~\textit{et al.}~\cite{zhan2017distinguishing} employed a CNN to distinguish clouds from snow in satellite imagery.
    A perceptual generative adversarial network (GAN) for single-image de-snowing was proposed in~\cite{wan2019perceptual}. Subsequent GAN-based methods introduced compositional~\cite{li2019composition} and two-stage architectures~\cite{jaw2020desnowgan} for more effective snow removal. Transformer-based architectures have also been proposed, incorporating global context~\cite{zhang2022desnowformer}, multi-scale projection~\cite{chen2023msp}, and context interaction~\cite{chen2022snowformer}. A deep invertible separation method was introduced in~\cite{quan2023image} for single image de-snowing. In~\cite{Lai_2025_CVPR}, a SnowMaster framework was proposed for real-world de-snowing using MLLM with multi-model feedback optimization.

\subsubsection{\textbf{Video De-snowing}} 
    Major de-snowing approaches are discussed below.
    \textit{\underline{Prior-based Methods:}} A depth prior-based stable tensor decomposition method for video snow removal was introduced in~\cite{li2024depth}, incorporating semantic and geometric priors. In~\cite{li2021video}, a saliency-guided approach using dual adaptive spatiotemporal filtering and guided filtering was proposed.
    \newline
    \textit{\underline{Learning-based Methods:}} \cite{xue2024rvdnet} introduced RVDNet, a two-stage network for real-world video de-snowing with domain adaptation, improving the performance of video snow removal. Video de-snowing remains underexplored, often addressed alongside other weather effects like rain or haze.
    
\subsection{Traffic Sign Interpretation}
Recent work has moved from isolated traffic sign detection toward \emph{traffic sign interpretation}, aiming to derive semantic driving instructions. To overcome the limitations of symbol- or text-based methods in modeling complex regulations, structured reasoning pipelines and traffic knowledge graphs were introduced in SignParser~\cite{10.1109/TITS.2024.3428619}, and the task was later reformulated as a vision--language problem for interpretable understanding. Yang \textit{et al.}~\cite{yang2023traffic,yang2024signeyetrafficsigninterpretation} formalized the Traffic Sign Interpretation (TSI) task through the SignEye framework, which integrates egocentric spatial reasoning and multi-task learning supported by the large-scale TSI-CN dataset. Similarly, Guo \textit{et al.}~\cite{guo2023visual} proposed Visual Traffic Knowledge Graph Generation (VTKGG) to model relations among roads, lanes, and traffic signs using hierarchical graph attention networks. Despite these advances, most interpretation methods assume clear and readable signs. In real-world transportation, sign text and symbols are often degraded by adverse weather, low illumination, motion blur, and lens contamination. Unlike general restoration, \emph{traffic sign text restoration} must preserve fine strokes and precise character geometry despite small size, perspective distortion, and reflective surfaces. Moreover, it should be evaluated by \emph{functional correctness} (e.g., OCR and rule extraction), since even minor artifacts can change sign semantics (e.g., speed limits or lane restrictions). Such errors can propagate to interpretation modules, causing incorrect reasoning and unsafe decisions. Thus, robust sign restoration under adverse conditions remains critical yet underexplored.

\section{Related Work: Multi-weather Restoration}
\label{sec: multi-weather restoration}
    This section discusses multi-weather (multi-task) image and video restoration approaches. Most of the existing methods are task-wise fine-tuned and evaluated across multiple weather (tasks) conditions. We discuss task-specific fine-tuned models evaluated on de-hazing, de-raining, and de-snowing tasks. 
    \vspace{-3mm}
    \subsection{Image Restoration}
        \noindent \underline{\textit{Prior-based Methods:}} \cite{wang2017hierarchical} introduced a hierarchical approach for rain and snow removal from single color images. \cite{dong2023framework} developed a joint framework for degraded image restoration and simultaneous localization processes.
        \newline
        \underline{\textit{Learning-based Methods:}} Chen \textit{et al.} \cite{chen2019gated} introduced leveraging gated context aggregation for haze and rain removal. A dual-tree complex wavelet fusion based approach is proposed in \cite{zang2020deraining} for rain and snow removal. Zaamir \textit{et al.} \cite{zamir2021multi} introduced a multi-stage network, that progressively acquires restoration functions for the degraded inputs. In \cite{kulkarni2022unified2}, a memory replay training strategy is adapted for multi-weather (\textit{haze, rain and snow}) image restoration. Zaamir \textit{et al.} \cite{zamir2022restormer} proposed restoration transformer by designing multi-head attention and feed-forward network for restoration. Wang \textit{et al.} \cite{wang2022uformer} presented Uformer, an effective and efficient Transformer-based encoder-decoder architecture for image restoration. zhou \textit{et al.} \cite{zhou2023fourmer} proposed a fourier spatial interaction modeling and Fourier channel evolution for image restoration. Gao \textit{et al.} \cite{gao2023frequency} proposed a frequency-oriented transformer excelling in weather-degraded image restoration. A transformer with grid-based feature fusion \cite{wang2024gridformer} and degradation-aware \cite{zhu2024multi} approaches are proposed for multi-weather image restoration. First semi-supervised learning framework based on vision-language model is proposed in \cite{xu2024towards}. Qin \textit{et al.} \cite{Qin_2025_CVPR} presented ResFlow: a image restoration framework that models the degradation process as a deterministic path using continuous normalizing flows. Kulkarni \textit{et al.} \cite{kulkarni2022wipernet} proposed WiperNet, a lightweight model for restoring images degraded by haze, rain, and snow with low computational cost. Kulkarni \textit{et al.} \cite{kulkarni2022unified} introduced an even smaller network (1.1M parameters) for rain and snow, but with limited generalization to extreme conditions. Patil \textit{et al.} \cite{patil2023multi} proposed a domain translation framework that generates weather-specific variants from a single input, but increases complexity, cost, and error propagation risk.
        
    \subsection{Video Restoration}
        \noindent \underline{\textit{Prior-based Methods:}}
        In~\cite{ren2017video} Matrix decomposition is proposed for video de-snowing and de-raining, using a weighted average approach. In \cite{zandersen2021nature} highlighted the role of nature-based solutions for climate adaptation, focusing on restoring environments affected by weather conditions. 
        \underline{\textit{Matrix Decomposition Based Methods:}} Kim \textit{et al.}\cite{kim2015video} proposed a video de-raining and de-snowing algorithm that leverages temporal correlation and low-rank matrix completion.
        \newline
        \underline{\textit{Learning-based Methods:}} A consolidated adversarial network for video de-raining and de-hazing task is proposed in~\cite{galshetwar2022consolidated}. In~\cite{patil2022video}, the authors emphasized meta-adaptation techniques for video de-hazing and de-raining under veiling effects in data-scarce scenarios. A dual-frame spatio-temporal feature modulation framework is proposed in~\cite{patil2022dual} to address degradation from diverse weather conditions.
        Above discussed methods achieve significant performance for multi-weather degraded image restoration. However, there are many challenging aspects where future work may rely on.
        \begin{itemize}
        \item Multi-weather restoration models must adaptively handle diverse real-world degradations, including varying rain and snow intensities and non-uniform haze.
         
        \item The model should minimize computational load; trainable parameters, inference time, model size, and FLOPs—for real-time multi-weather restoration.
        
        \item Despite progress in multi-weather image restoration, advancements in video restoration remain limited.
    \end{itemize}

\section{Related Work: All-in-One Image Restoration}
\label{sec: all-in-one}

    The emergence of all-in-one image restoration models represents a major advancement in handling multiple visual degradations, such as haze, noise, blur, low-light, rain, smog, and snow within a single unified framework. These models are trained once on a combined dataset and deployed across various adverse conditions without task-specific fine-tuning. This generalization capability makes them particularly valuable for intelligent transportation systems, where consistent, real-time visual clarity is crucial across diverse weather scenarios. With the scope of this survey, we focus on existing all-in-one approaches targeting key degradations, including haze, rain, noise, and snow. We categorize recent all-in-one methods based on their architectural principles.

\begin{table*}[t]
    \centering
    \scriptsize
    \caption{Overview of image dehazing datasets: the first column lists key metadata (venue, resolution, best PSNR/SSIM)}
    \resizebox{\linewidth}{!}{
    \begin{tabular}{m{2.9cm} m{1cm} m{17.6cm}} 
        \toprule
        \multicolumn{3}{c}{\textbf{Real Datasets}} \\
        \midrule
        \textbf{Name \& Metadata} & \textbf{Type} & \textbf{Dataset construction and insights} \\
        \midrule

        \makecell[l]{\textbf{Dense-Haze~\cite{Dense_Haze_2019}}, ICIP-19 \\ Resolution: 5456$\times$3632 \\ 17.55 / 0.67~\cite{Luo_2025_CVPR}} &  Real Dataset & \textbf{Samples: 33 image pairs.} Real haze was produced using professional machines (LSM1500 PRO 1500 W) to mimic atmospheric conditions, with images captured under consistent lighting (cloudy, morning/evening) and low wind (<3 km/h) for uniform haze. Identical settings and static scenes ensured accurate haze-free and hazy image pairs.\\
        \midrule

        \makecell[l]{\textbf{NH-HAZE~\cite{ancuti2020nh}}, CVPRW-20 \\ Resolution: 1600$\times$1200 \\ 29.46 / 0.890 \cite{ancuti2020nh}} & Real Dataset & \textbf{Samples: 55 image pairs.} Non-Homogeneous Haze dataset. It is a real-world outdoor images, each consisting of a hazy image and its corresponding haze-free ground truth. To simulate realistic haze conditions, the authors employed a professional haze generator that produces non-uniform haze distributions, closely mimicking real atmospheric scenarios. Images were captured under consistent lighting and environmental settings to ensure accurate pairing between hazy and haze-free images.\\ 
        \midrule

        \makecell[l]{\textbf{BeDDE \cite{8784729}}, ICME-19 \\ Resolution: 1643$\times$1200 \\ 0.9012/ 0.9725 (VI/RI) \cite{ling2024singleimagedehazingusing}} & Real Dataset & \textbf{Samples: 208 image pairs.} This is the first real-world dataset of foggy images paired with aligned clear counterparts, captured across diverse outdoor scenes. Each pair includes manually labeled masks for region-specific evaluation. Two new metrics are introduced: Visibility Index (VI) for visibility enhancement and Realness Index (RI) for perceived naturalness—offering both objective and subjective assessment of defogging performance.\\
        \midrule

        \makecell[l]{\textbf{Night-Haze \cite{Filin2022HazyID}}, DLCP-22 \\ Resolution: 6000$\times$4000 \\ 30.38/0.904 \cite{jin2024enhancingvisibilitynighttimehaze}} & Real Dataset & \textbf{Samples: 32 image pairs, Extended: 64 image pairs.} All images were captured indoors to maintain consistent conditions, the dataset includes two scenes—one with simple geometric objects, the other with complex, detailed objects and localized lighting. Each scene was imaged under four lighting and four haze levels, yielding 16 images per scene (32 total). The extended version, Night-Haze-Ext, offers 64 images with additional haze levels, scene variations, and includes depth and thermal data.\\
        \midrule

        \multicolumn{3}{c}{\textbf{Synthetic Datasets}} \\
        
        \midrule
        
         \makecell[l]{\textbf{RESIDE}\cite{li2019benchmarking}, TIP-19 \\ Resolution: 620$\times$460 \\ 36.39 / 0.988~\cite{Qin_Wang_Bai_Xie_Jia_2020}}
         & Synthetic Dataset & \resizebox{\linewidth}{!}{\begin{tabular}{m{2.5cm} m{15.5cm}}
        \makecell[l]{\textbf{Subsets \& Samples} \\ ITS: 13.9k image pairs \\ SOTS: 500 image pairs \\ HSTs: 20 image pairs \\ OTS: 72.1k image pairs \\ RTTS: 4.3k image pairs} & \vspace{3pt} Realistic Single Image Dehazing dataset (RESIDE). A large-scale synthetic training set, and two different sets designed for objective and subjective quality evaluations, respectively. This dataset includes both synthetic and real-world hazy images, divided into subsets like Indoor Training Set (ITS), Outdoor Training Set (OTS), Standard Testing Set (SOTS), Hybrid Subjective Testing Set (HSTS), and Real-world Task-driven Testing Set (RTTS). The synthetic images are generated using the atmospheric scattering model, combining clean images with depth information to simulate haze.
        \end{tabular}} \\
        \midrule
        
        \makecell[l]{\textbf{REVIDE \cite{REVIDE}}, CVPR-21 \\ Resolution: 2708$\times$1800 \\ 25.79 / 0.899 \cite{galshetwar2024cross}} & Synthetic Dataset & \textbf{Samples: 48 video pairs.} A real-world video dehazing dataset for supervised learning, captured using a robot arm, Sony ICLE 6000 camera, and haze machines to ensure precise alignment of hazy and haze-free video pairs. It features indoor scenes with realistic atmospheric scattering, offering high-quality data for training and evaluating video dehazing models.\\
        \midrule

         \makecell[l]{\textbf{HazeRD \cite{8296874}}, ICIP-17 \\ Resolution: $\sim$3000$\times$2448 \\ 18.55 / 0.85 \cite{liu21DMT-Net}} & Synthetic Dataset & \textbf{Samples: 14 image pairs.} The dataset contains 14 high-resolution (6–8 MP) haze-free RGB outdoor images, each paired with a depth map. Synthetic hazy versions are generated using the Koschmieder scattering model across five haze levels (visual ranges: 50m to 1000m), simulating varying atmospheric conditions based on scene geometry.\\
        \midrule
        
        \makecell[l]{\textbf{SOTS \cite{li2019benchmarking}}, TIP-19 \\ Resolution: 620$\times$460 \\ 39.42 / 0.996 \cite{10571568}} & Synthetic Dataset & \textbf{Samples: indoor 500 and 500 outdoor.} SOTS evaluates single-image dehazing under controlled settings with two subsets: SOTS-Indoor and SOTS-Outdoor, both containing synthetic hazy images. Haze-free images with estimated depth maps were used to generate realistic haze via the atmospheric scattering model.\\
        \midrule
        
        \makecell[l]{\textbf{DAVIS-2016 \cite{patil2022dual}}, CVIP-21 \\ Resolution: 256$\times$256 \\ 22.67 / 0.879 \cite{galshetwar2024cross}} & Synthetic Dataset & \textbf{Samples: 50 video pairs.} Synthetic Outdoor video de-hazing dataset that is generated synthetically and depth maps of each video frame of DAVIS-16 video dataset. Depth map of each respective frame in a video is estimated using the approach proposed in \cite{ranftl2020towards}. The attenuation coefficient  = 2 and the atmospheric light value A = (0.8, 0.8, 0.8) were taken into account while generating the synthetic dataset.\\
        \midrule
        
        \makecell[l]{\textbf{NYU-Depth \cite{6130298}}, ICCVW-11 \\ Resolution: 256$\times$256 \\ 23.81 / 0.897 \cite{galshetwar2024cross}} \vspace{-1mm} & Synthetic Dataset & \textbf{Samples: 45 video pairs.} Synthetic indoor dataset. It contains 45 videos divided into training (25 videos/ 28,222 frames) and testing (20 videos/ 7528 frames) videos. Depth maps are used to generate the synthetic hazy videos.\\
        \midrule
        
        \makecell[l]{\textbf{D-Hazy \cite{ancuti2016dhazy}}, IEEE CIP-16 \\ Resolution: [640$\times$480, \\ 1024$\times$768] \\ 28.25 / 0.937 \cite{10414993}} & Synthetic Dataset & \textbf{Samples: 22 image pairs}. A high-quality synthetic image dataset is generated synthetically. Ground-truth clear images and depth maps were taken from the Middlebury stereo dataset. The Middlebury dataset provides high-quality stereo images along with accurate depth maps. These were used to simulate realistic haze conditions by varying parameters like Atmospheric light A, Scattering coefficient $\beta$.\\
        \midrule
        
        \vspace{-4mm}\makecell[l]{\textbf{I-Hazy \cite{Ancuti2018}}, ACIVS-18 \\ Resolution: 2833$\times$4657 \\ 22.44 / 0.887 \cite{10571568}} & Synthetic Dataset & \textbf{Samples: 35 image pairs.} A total of 35 indoor scenes with varied household objects and surface properties were set up, each including a Macbeth ColorChecker for color calibration. For each scene, a haze-free image was captured under controlled lighting, followed by a hazy image after introducing real atmospheric-like haze using two fog machines (LSM1500 PRO 1500 W) and a fan to ensure even distribution. Both images were taken under identical lighting conditions.\\
        \midrule
        
        \vspace{-4mm}\makecell[l]{\textbf{V-Hazy \cite{dudhane2020varicolored}}, CVPR-20 \\ Resolution: [640$\times$480, \\ 1024$\times$768] } & Synthetic Dataset & \textbf{Samples: 35 image pairs.} The author created a synthetic varicolored hazy image dataset by using the channel-wise spatial mean of real-world hazy images as atmospheric light, preserving the haze color. Hazy images were categorized into grayish, orange/yellow (smog), bluish, and other variants. Synthetic images were generated with haze densities defined by $\beta = {1, 3, 5}$. Additionally, color-balanced versions were created using $A = (0.8, 0.8, 0.8)$ and the same $\beta$ values.\\
         
        \bottomrule
    \end{tabular}}
    \label{TAB:dehaze_dataset}
\end{table*}

\begin{table*}[t]
    \centering
    \scriptsize
    \caption{Overview of image deraining datasets: the first column lists key metadata (venue, resolution, best PSNR/SSIM)}
    \resizebox{\linewidth}{!}{
    \begin{tabular}{m{3.4cm} m{1.4cm} m{15.7cm}}
        \toprule   

        \multicolumn{3}{c}{\textbf{Real Datasets}} \\
        \midrule
        
        \textbf{Name \& Metadata} & \textbf{Type} & \textbf{Dataset description and insights} \\
        \midrule

        \makecell[l]{\textbf{Rain12 \cite{Yang2016DeepJR}}, CVPR-16 \\ Resolution: 512$\times$512 \\ 36.69 / 0.962 \cite{8953349}} \vspace{-1mm} & Real Dataset & \textbf{Samples: 12 rainy images.} The authors proposed a realistic rain simulation model combining Rain Streaks (with varied shapes and directions) and Rain Accumulation (atmospheric veils mimicking mist/fog). A key innovation is the rain-streak binary map, labeling each pixel for streak presence to separate rain-affected areas from the background. The resulting dataset includes: (1) synthetic rainy images, (2) corresponding clean images, and (3) pixel-level binary maps of rain streaks.\\
        \midrule

        \makecell[l]{\textbf{Real-world \cite{wang2019spatial}}, CVPR-19 \\ Resolution: 1000$\times$1000 \\ 36.55 / 0.962 \cite{Wang_2023_CVPR}} \vspace{-1mm} & Real Dataset & \textbf{Samples: 29500 image pairs.} Spatially Aligned Paired Data is a large-scale real-world rainy image dataset captured using professional cameras. Each image pair—one with rain and one without—was taken from nearly identical viewpoints using tripods and remote shutters. Efforts were made to match illumination conditions, and frames with moving objects were manually selected to avoid mismatches. The resulting pairs have minimal misalignment, making them ideal for supervised learning.\\
        \midrule

        \multicolumn{3}{c}{\textbf{Synthetic Datasets}} \\
        \midrule
        
        \vspace{-1mm} \makecell[l]{\textbf{RID \cite{8953352}}, CVPR-19 \\ Resolution: 512$\times$512 \\ 7.625 / 7.492 / \\ 23.93 / 34.61 \cite{patil2023multi}} \vspace{-1mm} & Synthetic Dataset & \textbf{Samples: Indoor 16,200 and Outdoor 10,500.} The authors introduce NYU-Rain, a synthetic rain dataset built from NYU-Depthv2 images by rendering rain streaks and accumulation effects using depth information, including veiling and blur (see Algorithm 1). It comprises 16,200 samples, with 13,500 for training. They also create Outdoor-Rain, an outdoor rain dataset generated using depth estimated via state-of-the-art single-image depth methods, containing 9,000 training and 1,500 validation samples.\\
        \midrule

        \vspace{-2mm}\makecell[l]{\textbf{RTTS}\cite{8954466}, CVPR-19 \\ Resolution: 620$\times$460 \\ 24.76 / 42.04~\cite{patil2023multi}} & Synthetic Dataset & \textbf{Samples: 13900 image pairs}. Realistic multi-purpose single image deraining dataset. Synthetic rain streak images created by overlaying computer-generated rain streaks onto clean images. Synthetic raindrop images generated by simulating raindrops on camera lenses, using a binary mask to define raindrop regions. Synthetic images that combine rain streaks with atmospheric scattering effects to simulate mist, using a model that includes transmission maps $\&$ atmospheric light. \\
        \midrule
        
        \makecell[l]{\textbf{RainCityscapes \cite{hu2019depth}}, CVPR-19 \\ Resolution: 2048$\times$1024 \\ 35.82 / 0.987 \cite{PARK2022116701}} \vspace{-1mm} & Synthetic Dataset & \textbf{Samples: $\sim$10,000 image pairs.}  The RainCityscapes dataset was created by adding synthetic rain to Cityscapes images using a depth-guided, physically-inspired model. Rain streaks vary in length (based on speed and exposure), direction (wind-influenced), and transparency (more opaque at shallow depths). Depth maps enable realistic effects—closer objects show sharper streaks, while distant areas appear blurred. A veiling effect, simulating light scattering like fog, is added using depth-based exponential decay.\\
        \midrule

        \vspace{-3mm}\makecell[l]{\textbf{Rain800 \cite{zhang2019image}}, CVPR-19 \\ Resolution: 512$\times$512 \\ 32.00 / 0.923 \cite{zhang2019image}} & Synthetic Dataset & \textbf{Samples: 800 image pairs.} The authors utilized clean images from publicly available datasets as the foundation for creating synthetic rainy images. Rain streaks were algorithmically added to the clean images to simulate various rain conditions. This process involved controlling parameters such as streak orientation, density, and intensity to mimic real-world rain patterns. Each synthetic rainy image was paired with its original clean counterpart. \\
        \midrule
        
        \vspace{-2mm} \makecell[l]{\textbf{Rain100H \cite{7780668}}, CVPR-16 \\ Resolution: 480$\times$320 \\ 34.56 / 0.941 \cite{10.5555/3692070.3694465}} \vspace{-1mm} & Synthetic Dataset & \textbf{Samples: 1900 image pairs.} This dataset contains high-quality synthetic rainy images with corresponding clean ground truth. Clean images from public datasets were used to generate diverse scenes, and rain was simulated using a Physical Rain Model and Layer-Based Separation. The rain model applied Gaussian-distributed streaks with varied length, width, and direction, along with motion blur to mimic real rain. The model assumes that the observed rainy image I is the sum of the background layer B and the rain streak layer R: $I=B+R$. This layered approach helped the network learn how to remove rain while preserving background details. Multiple rainy variants were created per clean image to simulate diverse conditions.\\
        \midrule
        
        \vspace{-2mm}\makecell[l]{\textbf{DID-Data \cite{8099669}}, CVPR-17 \\ Resolution: 512$\times$512 \\ 35.66 / 0.967 \cite{Yamashita_2024_ACCV}} & Synthetic Dataset & \textbf{Samples: 13,200 image pairs.} Synthetic rainy images were created by adding artificially rendered rain streaks of varying intensity, direction, and appearance to high-quality clear images sourced online. Tools like Photoshop were used for realism. Each clean image was paired with multiple rainy variants, forming a supervised dataset of (rainy, clean) image pairs. \\ 
        \midrule
        
        \vspace{-2mm} \makecell[l]{\textbf{DIDMDN-Data \cite{8578177}}, CVPR-18 \\ Resolution: 512$\times$512 \\ 30.57 / 0.8719 \cite{10.1145/3577530.3577543}} \vspace{-1mm} & Synthetic Dataset & \textbf{Samples: 13,200.} Synthetic rainy images with light, medium, and heavy rain were created by adding simulated rain to clean images from datasets like BSD500 and UCID. Rain streaks of varying densities were generated with diverse orientations, sizes, and intensities, followed by motion blur for realism. These rain layers were blended with clean images, and each output was labeled by rain density. This enabled training a density-aware network using a large set of synthetic (rainy, clean) image pairs.\\
        \midrule

        \makecell[l]{\textbf{R200H and R200L \cite{yang2017deep}}, CVPR-19 \\ Resolution: 512$\times$512 \\ 32.99 / 0.940 \\ 41.81 / 0.990 \cite{Yamashita_2024_ACCV}} \vspace{-1mm} & Synthetic Dataset & \textbf{Samples: 2000 image pairs.} Synthetic rainy images were generated by overlaying varied rain streaks (in angle, shape, transparency, and motion blur) onto clean outdoor backgrounds sourced from public datasets. Multiple rain layers simulated light and heavy rain. The rainy image formation followed $O=(B+R)\times T+A(1-T)$, where $O$ is the observed image, $B$ the clean background, $R$ rain streaks, $T$ transmission, and $A$ atmospheric light. Rain masks were also created to aid supervised training.\\
        \midrule

        \makecell[l]{\textbf{AGAN-Data \cite{qian2018attentive}}, CVPR-18 \\ Resolution: 512$\times$512 \\ 32.45 / 0.937 \cite{zhou2024adapt}} \vspace{-1mm} & Synthetic Dataset & \textbf{Samples: 1119 image pairs.} A synthetic dataset for raindrop removal was created using image pairs of identical scenes—one with raindrops and one clean—captured through two identical glass slabs (one sprayed with water). This setup avoids misalignment caused by refraction. Camera motion and environmental factors were controlled to ensure consistency. Images were captured using Sony A6000 and Canon EOS 60, with 3 mm thick glass placed 2–5 cm from the lens to vary raindrop patterns and minimize reflections.\\ 
        \midrule
        
        \makecell[l]{\textbf{ORD \cite{6353522}}, IEEE SPL-13 \\ Resolution: 96$\times$96 \\ 32.05 / 0.952 \cite{patil2023multi}}  & Synthetic Dataset & \textbf{Samples: 9750 image ($~$ 250,000 patches of size 96×96 pixels).} provided the degraded images with rain and fog having veiling effect degradation. \\

        \bottomrule
    \end{tabular}}
    \label{TAB:derain_dataset}
\end{table*}

\begin{table*}[t]
    \centering
    \scriptsize
    \caption{Overview of image desnowing datasets: the first column lists key metadata (venue, resolution, best PSNR/SSIM)}
    \resizebox{\linewidth}{!}{
    \begin{tabular}{m{3cm} m{1.3cm} m{16.2cm}}
        \toprule   

        \multicolumn{3}{c}{\textbf{Synthetic and Real Datasets}} \\
        \midrule

        \textbf{Name \& Metadata} & \textbf{Type} & \textbf{Dataset description and insights} \\
        \midrule
    
        \makecell[l]{\textbf{SRRS \cite{Chen2020}}, ECCV-20 \\ Resolution: 1920$\times$1080 \\ 32.39 / 0.98 \cite{10571568}} \vspace{-1mm} & Synthetic and Real Dataset & \textbf{Samples: Synthetic 50 video pairs (500 frames per video), Real-World: 5 videos (500 frames per video)} Captures video sequences of snowfall to reflect temporal snow dynamics—such as snowflake motion and veiling effects—not visible in static images. Clean videos were used to generate the dataset, with snow particles rendered using tools like Photoshop. Simulations vary in particle size, transparency, motion, and density (light to heavy). Real scenes are also included.\\
        \midrule
        
        \makecell[l]{\textbf{Snow100K \cite{liu2018desnownet}}, TIP-18 \\ Resolution: 640$\times$640 \\ 33.92 / 0.96 \cite{10571568}} \vspace{-1mm} & Synthetic $\&$ Real set & \textbf{Samples: 100k synthesized snowy image pairs and 1,329 realistic snowy images.} To build a diverse snow image dataset, the authors synthetically added snow to clean images from sources like ImageNet and COCO, treating the originals as ground truth. Snowflakes—varying in size, shape, and transparency—were generated based on realistic distribution models and overlaid either randomly or in patterns. Each synthetic image includes a snow mask marking snowflake locations, enabling precise evaluation. Both opaque and translucent snow effects were simulated. Real snowy images were also collected (e.g., from Flickr), with manually annotated snow masks. The dataset is split into three subsets by snowflake size: Snow100K-S (small), Snow100K-M (small+medium), and Snow100K-L (small+medium+large), each with ~33K images.\\
        \midrule

        \multicolumn{3}{c}{\textbf{Synthetic Datasets}} \\
        \midrule

        \makecell[l]{\textbf{RVSD \cite{Chen_2023_ICCV}}, ICCV-23 \\ Resolution: 1920$\times$1080 \\ 26.02 / 0.923 \cite{Ghasemabadi2024LearningTC}}  & Synthetic Dataset & \textbf{Samples: 110 video pairs.} The authors created a comprehensive dataset for training and evaluating video snow removal algorithms using Unreal Engine 5 and augmentation techniques to simulate realistic snow and haze across diverse scenes and conditions.\\
        \midrule

        \makecell[l]{\textbf{SnowCityScapes \cite{9515587}}, TIP-21 \\ Resolution: 512$\times$256 \\  38.60/0.9822 \cite{zhang2021deep}} & Synthetic Dataset & \textbf{Samples: 15,000 image pairs.} Based on the Cityscapes dataset, known for its high-quality urban street scenes. Utilized Adobe Photoshop to overlay synthetic snow onto the clean images. Encompasses three snow conditions: light, medium, and heavy snow. Comprises paired images: synthetic snowy images and their corresponding clean images. Maintains consistency with the original Cityscapes dataset in terms of image size and scene content. \\
        \midrule
        
        \vspace{-2mm}\makecell[l]{\textbf{SnowKITTI \cite{9515587}}, TIP-21 \\ Resolution: 1242$\times$375 \\ 38.96 / 0.99 \cite{chen2022snowformer}} & Synthetic Dataset & \textbf{Samples: 1,167 image pairs.} Derived from the KITTI 2012 dataset, which comprises real-world driving scenes. Utilized Adobe Photoshop to overlay synthetic snow onto the clean images. Simulated three snow conditions: light, medium, and heavy snow. It includes both training and testing sets. Each set contains image pairs: the synthetic snowy image and its corresponding clean image. Each image was augmented to simulate 3 snow conditions: light, medium, and heavy snow. \\
        \midrule
        
        \makecell[l]{\textbf{CSD \cite{chen2021all}}, TCSVT-21 \\ Resolution: 640$\times$480 \\ 32.95 / 0.942 \cite{patil2023multi}} \vspace{-1mm} & Synthetic Dataset & \textbf{Samples: 110 pairs of videos.} CSD combines synthetic and real snowy images for training and evaluation. Snow-free backgrounds from datasets like ImageNet and COCO were overlaid with simulated snowflakes of varying size, shape, opacity, and motion blur using layered alpha blending to create light, medium, and heavy snowfalls. Real-world snowy images were also sourced from platforms like Flickr and Google, selected for clear snowfall and diverse conditions.\\

        \bottomrule
    \end{tabular}}
    \label{TAB:desnow_dataset}
\end{table*}

\subsection{Prompt-Based and Adaptive Architectures}
Prompt-based methods have emerged as a flexible approach to guide restoration processes across diverse degradation types. For instance, PromptIR~\cite{potlapalli2023promptir} introduces degradation-specific prompts that modulate the restoration network, enabling effective handling of de-hazing, denoising, and de-raining tasks within a single model.
    Building upon this, Adaptive Blind All-in-One Restoration (ABAIR)~\cite{serrano2024adaptive} incorporates a segmentation head to estimate per-pixel degradation types, facilitating the model's adaptability to unseen degradations. By employing low-rank adapters, ABAIR efficiently integrates new degradation types with minimal parameter updates, enhancing its applicability in dynamic environments.
    Li \textit{et al.} \cite{10697214} proposed a U-shaped convolutional network designed to restore images degraded by various adverse weather conditions. It utilizes traditional 2D convolutions for feature extraction and incorporates a prompt generation module to create weather-specific prompts that guide the decoding process. Additionally, frequency separation via wavelet pooling is employed to enhance high-fidelity restoration.
\subsection{Frequency and Feature Perturbation Techniques}
Addressing the challenge of task interference in multi-degradation scenarios, AdaIR~\cite{cui2024adair} leverages frequency mining and modulation to adaptively reconstruct images. By accentuating informative frequency subbands corresponding to specific degradations, AdaIR achieves state-of-the-art (SOTA) performance in tasks including denoising and de-hazing.
    Similarly, Degradation-aware Feature Perturbations (DFPIR) \cite{Tian_2025_CVPR}, employing channel-wise and attention-wise perturbations to align feature representations with the shared parameter space. This strategy mitigates task interference, enhancing the model's capability to handle multiple degradations such as noise and haze effectively.
    
\subsection{State Space Models and Diffusion-Based Approaches}
    DPMambaIR~\cite{liu2025dpmambair} combines degradation-aware prompting with high-frequency enhancement for fine-grained restoration of snow, haze, and noise. Diffusion-based methods~\cite{article, ozdenizci2023restoring} model weather distributions via latent mapping and conditional transformers, but are computationally heavy and struggle with extreme or unseen conditions. AutoDIR~\cite{jiang2024autodir} uses latent diffusion for adaptive restoration, offering stronger real-world generalization.
    
\subsection{Transformer-Based and Mixture-of-Experts Models} Transformer-based models like TransWeather~\cite{valanarasu2022transweather} employ intra-patch attention and learnable weather-type embeddings within a unified encoder-decoder framework to adaptively restore images degraded by haze, snow, and other conditions. A vision transformer in~\cite{10767188} leverages contrastive learning to extract distortion-aware features for multi-weather restoration but shows performance drops when tested on unseen weather types. Adaptive sparse transformer in~\cite{zhou2024adapt} leverages attentive feature refinement to mitigate noisy interactions for image restoration. The Weather-aware Multi-scale Mixture-of-Experts~\cite{luo2023wm} dynamically routes inputs to specialized experts based on weather conditions, improving restoration under complex, mixed degradations such as snow and haze. Multi-branch linear transformer in~\cite{jin2025mb} leverages Taylor formula for image restoration.

\subsection{Architectural Search Approach} 
    Li~\textit{et al.}~\cite{li2020all} proposed a unified deep learning framework for multi-weather image restoration using neural architecture search (NAS), which automatically identifies optimal architectures for handling rain, snow, and haze, thereby improving generalization across diverse weather degradations.
    
\subsection{Multi-Modality Approach} 
    Siddiqua~\textit{et al.}~\cite{siddiqua2021multi} used a conditional GAN with multi-modal inputs \textit{i.e.} RGB images and contextual information to restore the image. While effective, the model incurs high computational costs during both training and inference. A multi-domain attention-based conditional adversarial network is proposed in \cite{siddiqua2023macgan} for all-in-one image restoration.

\subsection{Language-driven Approach}
    Ai \textit{et al.} \cite{ai2024multimodal} proposed MPerceiver, which uses multimodal prompt learning with Stable Diffusion priors by combining textual prompts for global guidance and visual prompts for multi-scale refinement. Conde \textit{et al.} \cite{conde2024high} leveraged real human-written instructions for multi-task restoration. Yang \textit{et al.} \cite{yang2024language} introduced a language-driven all-in-one adverse weather removal method using vision-language priors for flexible, user-guided restoration across diverse conditions. However, its Mixture-of-Experts (MoE) expert selection increases training and tuning complexity.

\subsection{Knowledge Distillation Approach}
    Chen~\textit{et al.}~\cite{chen2022learning} proposed a unified model for removing haze, snow, and rain using a single set of pretrained weights. It uses a two-stage process: Knowledge Collation transfers expertise from multiple teachers, and Knowledge Examination refines the student via multi-contrastive regularization. However, it adds significant training complexity and overhead.
    
\subsection{Weather-General and Weather-Specific Approach}
    Zhu \textit{et al.} \cite{10204275} proposed a two-stage strategy: Weather-General Feature Learning for coarse restoration across diverse degradations, followed by Weather-Specific Feature Learning that adaptively expands parameters to model each weather type. While effective, this limits scalability in highly dynamic real-world conditions. \textit{Unknown Corruption Approach:} Li \textit{et al.} \cite{9879292} proposed AirNet, which uses a contrastive degraded encoder to learn latent degradation representations and a degradation-guided restoration network to recover clean images from unknown corruptions.

\subsection{Agentic AI Methods:}
Recent all-in-one image restoration methods increasingly adopt agentic and generative paradigms to handle mixed degradations. Zhu \textit{et al.}~\cite{zhu2024intelligent} decomposed compound restoration into modular tasks coordinated by specialized agents. Liu \textit{et al.} \cite{liu2025fapeirfrequencyawareplanningexecution} integrated semantic reasoning with frequency-aware diffusion using multimodal language models for restoration planning and zero-shot generalization. Rajagopalan \textit{et al.} \cite{rajagopalan2025restorevarvisualautoregressivegeneration} proposed visual autoregressive generation for fast coarse-to-fine restoration, while Jiang \textit{et al.} \cite{jiang2025cataircontenttaskawareallinone,jiang2025multiagentimagerestoration} introduced content-aware and multi-agent frameworks for structured recovery of unknown degradations. Li \textit{et al.} \cite{li2025hybridagentsimagerestoration} further combined fast, slow, and feedback agents to enhance robustness under mixed distortions. Collectively, these works highlight a shift toward planning-driven, agent-based, and efficient generative frameworks for scalable image restoration.

\begin{table*}[t]
\centering
\caption{Commonly used loss functions in image/video restoration tasks such as dehazing, deraining, and desnowing.}
\label{tab:loss_functions}
\resizebox{\linewidth}{!}{
\begin{tabular}{p{2.5cm}p{6cm}p{9cm}}
\toprule
\textbf{Loss Function} & \textbf{Mathematical Equation} & \textbf{Description and Usage} \\
\midrule

\textbf{Image Similarity Loss (PSNR-based)} & 
$L_{\text{psnr}} = \frac{10}{\log_{10}} \cdot \frac{1}{B} \sum_{b=0}^{B-1} \log(\| I_{\text{rst}} - I_{\text{gt}} \|_2 + \epsilon)$ & 
Measures pixel-wise quality via log-MSE; higher PSNR reflects better perceptual similarity and aids in tracking restoration tasks training.\\
\midrule

\textbf{Weather Classification Loss} & 
$L_{\text{cls}} = -\frac{1}{B} \sum_{b=1}^{B} \sum_{c=1}^{M} y_{bc} \log(p_{bc})$ \newline 
$y_{bc} \in \{0,1\}$: indicator if class $c$ is the true label for sample $b$ & 
A cross-entropy loss applied for classifying weather types (e.g., haze, rain, snow). Enables multi-task learning in restoration networks where weather condition labels assist in accurate image enhancement. \\
\midrule

\textbf{Reconstruction and Restoration Losses} & 
$L_{\text{rec}} = \| X' - X \|_1$, \quad 
$L_{\text{res}} = \| Y' - Y \|_1$, \newline 
$L_{\text{acc}} = \| \text{Model}(Y' - X) - Y \|_1$ & 
$X$: degraded image, $Y$: ground truth, $X'$, $Y'$: reconstructed outputs. These losses ensure fidelity at both representation and output level. Commonly used in encoder-decoder setups to guide accurate reconstruction. \\
\midrule

\textbf{Charbonnier Loss} & 
$L_{\text{char}} = \sqrt{ \| I^c - \hat{I} \|^2 + \varepsilon^2 }$, \quad $\varepsilon = 10^{-4}$ & 
A smooth, robust variant of L2 loss, commonly used in image restoration for preserving sharp details and ensuring training stability.\\
\midrule

\textbf{Edge Loss} & 
$L_{\text{edge}} = \sqrt{ \| \nabla I^c - \nabla \hat{I} \|^2 + \varepsilon^2 }$ & 
Encourages edge consistency using image gradients (Laplacian/Sobel), aiding fine-detail and texture restoration.\\
\midrule

\textbf{Mean Squared Error (MSE)} & 
$\mathcal{L}_{\text{MSE}} = \frac{1}{N} \sum_{i=1}^{N} (x_i - \hat{x}_i)^2$ & 
Penalizes squared pixel differences; standard for regression. Enables smooth restoration but can cause blurriness when used alone.\\
\midrule

\textbf{L1 Loss (MAE)} & 
$\mathcal{L}_{\text{L1}} = \frac{1}{N} \sum_{i=1}^{N} |x_i - \hat{x}_i|$ & 
Less sensitive to outliers than MSE, L1 loss preserves edges and yields sharper outputs by promoting pixel-wise sparsity. Commonly used in image restoration.\\
\bottomrule
\end{tabular}}
\end{table*}

\begin{table*}[t]
\centering

\caption{Common reference and no-reference evaluation metrics for image/video quality and generative model assessment.}
\label{TAB: parameter_table}
\resizebox{\linewidth}{!}{
\begin{tabular}{>{\bfseries}p{2.5cm} p{6cm} p{9cm}}
\toprule
\textbf{Metric} & \textbf{Mathematical equation} & \textbf{Description and usage} \\
\midrule
SSIM 
& 
$\text{SSIM}(x, y) = \frac{(2\mu_x\mu_y + C_1)(2\sigma_{xy} + C_2)}{(\mu_x^2 + \mu_y^2 + C_1)(\sigma_x^2 + \sigma_y^2 + C_2)}$ 
& 
Measures structural similarity between two images by combining luminance, contrast, and structural comparisons. Value ranges from \(-1\) to \(1\), with \(1\) indicating perfect similarity. \\
\midrule
PSNR 
& 
$\text{PSNR} = 10 \cdot \log_{10}\left( \frac{\text{MAX}^2}{\text{MSE}} \right)$ 
& 
Measures the ratio of peak signal to noise power; higher PSNR indicates better quality.\\
\midrule
LPIPS  
& 
(No closed-form; computed using pretrained deep networks)
& 
Measures perceptual similarity via deep features; lower LPIPS implies better similarity. Widely used in generative models.\\
\midrule
FID 
& 
$\text{FID} = \|\mu_r - \mu_g\|^2 + \text{Tr}\left(\Sigma_r + \Sigma_g - 2(\Sigma_r \Sigma_g)^{1/2}\right)$ 
& 
Compares real vs. generated features using InceptionNet; lower scores imply better generation. Sensitive to mode collapse.\\
\midrule
NIQE
& 
$\text{NIQE}(x) = (\mu_x - \mu_n)^T(\Sigma_x + \Sigma_n)^{-1}(\mu_x - \mu_n)$
& 
Quantifies naturalness by comparing image features to natural scene stats; lower NIQE scores indicate better perceptual quality.\\
\midrule
BRISQUE 
& 
(Model trained on NSS features and SVM regression)
& 
Extracts spatial scene statistics \& uses an SVM trained on subjective scores to predict quality. Low score better quality.\\
\midrule
Entropy 
& 
$H(I) = -\sum_{i=0}^{255} p(i) \log_2 p(i)$
& 
Measures image texture complexity; higher entropy may indicate more detail but doesn't always reflect perceptual quality.\\
\midrule
PIQ 
& 
(Deep feature-based metric, no analytical formula)
& 
Estimates perceptual quality using deep features \& learned weights, combining cues like sharpness, contrast, and texture. Lower scores indicate better quality.\\

\bottomrule
\end{tabular}}
\end{table*}

\section{Datasets for Image and Video Restoration}
\label{sec:III}
    In this Section, we have discussed the benchmark image and video datasets utilized to compare the current SOTA approaches for image/video de-hazing, de-raining and de-snowing tasks. The datasets are broadly classified into synthetic and real-world datasets. The TABLE \ref{TAB:dehaze_dataset}, \ref{TAB:derain_dataset} and \ref{TAB:desnow_dataset} provides a detailed overview of benchmark datasets used for respective application. Each row corresponds to a specific dataset, the first column includes metadata such as spatial resolution, publication venue, and best reported PSNR/SSIM values, while second column describes the dataset type (synthetic or real), and third column gives the insights into the construction methodology of each dataset, such as whether the images were synthetically generated, collected from real-world scenes, or created using paired or unpaired data. This comparison highlights diverse dataset designs and evaluation standards across weather degradations, emphasizing the need for realistic, high-quality datasets to benchmark and develop robust restoration methods.


\section{Loss Functions}
\label{sec:IV}
   Here, we have discussed the various existing loss-functions. Table~\ref{tab:loss_functions} presents a comprehensive summary of various loss functions commonly employed in image/video restoration tasks such as de-hazing, de-raining, and de-snowing. The first column lists the names of the loss functions, the second column provides their corresponding mathematical formulations, and the third column offers descriptions of their roles and applications in restoration models. These loss functions ranging from basic pixel-wise losses like L1 and L2 to more advanced perceptual, adversarial, and structural losses. These losses are crucial in guiding models to produce visually and quantitatively improved results. The descriptions highlight how each loss function contributes differently to model performance, such as improving edge sharpness, preserving texture details, or enhancing perceptual similarity. This structured presentation aids in understanding the trade-offs and suitability of loss functions for types of weather degradation scenarios.

\begin{table}[t]
\begin{center}
\caption{Subjective result Analysis on Real-world De-hazing (RTTS~\cite{li2019benchmarking}), De-raining with Veil (RID~\cite{8953352}) $\&$ in terms of Average NIQE, Entropy (ENT), Brisque (BRQ) $\&$ PIQE (PIQ). ($\downarrow$) lower is better, ($\uparrow$) higher is better.}
\resizebox{\linewidth}{!}{
\begin{tabular}{l|l|cccccccc}
\toprule
  \multirow{2}{*}{Dataset} & Evaluation  & \multicolumn{6}{c}{Methods} \\
  & Parameter  & UMVR \cite{kulkarni2022unified} & KD \cite{9879902} & TransWeather \cite{valanarasu2022transweather} & Diffusion \cite{ozdenizci2023restoring} & WGWS \cite{10204275} & DTMIR \cite{patil2023multi} & \\

    &  & TMM-22 & CVPR-22 & CVPR-22 & TPAMI-23 & CVPR-23 & ICCV-23 \\
  
\midrule
    &  NIQE ($\downarrow $) & 5.009 & 4.996 & 5.703 & 5.315 & 6.199 & \textbf{4.859} \\
   RTTS  &  ENT ($\uparrow $) & 7.221 & 7.297 & 7.263 & 7.115 & 7.064 & \textbf{7.505} \\
     &  BRQ ($\downarrow $) & 28.625 & 26.837 & 29.874 & 29.897 & 35.715 & \textbf{24.761} \\
     &  PIQ ($\downarrow $) & 42.749 & 45.561 & 43.784 & 50.386 & 55.296 & \textbf{42.037} \\
 \hline
     &  NIQE ($\downarrow $) & \textbf{6.604} & 6.943 & 7.496 & 7.103 & 6.813 & 7.625 \\
 RID &  ENT ($\uparrow $)& 7.486 & 7.459 & 7.393 & 7.401 & 7.348 & \textbf{7.492} \\
     &  BRQ ($\downarrow $) & 24.296 & 24.841 & 24.165 & 24.021 & 24.862 & \textbf{23.931} \\
     &  PIQ ($\downarrow $) & \textbf{33.289} & 38.398 & 39.141 & 36.651 & 35.094 & 34.614 \\

\bottomrule
\end{tabular}}
\label{TAB:NonRef}
\end{center}
\vspace{-4mm}
\end{table}


\section{Experimental Results and Discussion}

\begin{table*}[t]
  \centering
  \resizebox{\textwidth}{!}{
  \begin{tabular}{ccc}
  
  \begin{minipage}{0.25\textwidth}
    \centering
    \caption{Results of existing methods on Dense-Haze \cite{ancuti2019dense} dataset for real haze removal.}
    \resizebox{\linewidth}{!}{
    \begin{tabular}{l|c|c}
    \toprule
    \multicolumn{1}{c|}{\multirow{2}{*}{Method}} & \multicolumn{2}{c}{Dense-Haze}\\
  
         & PSNR & SSIM\\
    \midrule
    RIDCP \cite{wu2023ridcp} & 8.09 & 0.42 \\
    DCP \cite{5567108} & 10.06 & 0.39 \\
    SGID \cite{bai2022self} & 13.09 & 0.52 \\
    D4 \cite{yang2022self} & 13.12 & 0.53 \\
    AOD-Net \cite{li2017aod} & 13.14 & 0.41 \\
    GridDehazeNet \cite{liu2019griddehazenet} & 13.31 &  0.37 \\
    DA-Dehaze \cite{shao2020domain} & 13.98 & 0.37 \\
    FFA \cite{Qin_Wang_Bai_Xie_Jia_2020} & 14.39 &  0.45\\
    AECR-Net \cite{wu2021contrastive} & 15.80 &  0.47 \\
    DFormer \cite{song2023vision} & 16.29 & 0.51 \\
    DeHamer \cite{guo2022image} & 16.62 & 0.56 \\
    MBTFormer-B \cite{qiu2023mb} & 16.66 & 0.56 \\
    \midrule
    Uformer \cite{wang2022uformer} & 15.22 & 0.43 \\
    Restormer \cite{zamir2022restormer} & 15.78 & 0.55 \\
    Fourmer \cite{zhou2023fourmer} & 15.95 & 0.49 \\
    ResFlow \cite{Qin_2025_CVPR} & 17.12 & 0.59 \\
    \midrule
    AST-B \cite{zhou2024adapt} & 17.27 & 0.57 \\
    \textbf{Defusion \cite{Luo_2025_CVPR}} & \textbf{17.55} & \textbf{0.67} \\
    \bottomrule
    \end{tabular}}

    \label{TAB:ref6}
    \end{minipage}

  &
    \begin{minipage}{0.3\textwidth}
    \centering
    \captionof{table}{Single task evaluation: Video dehazing comparison on REVIDE dataset \cite{REVIDE}}
    \begin{tabular}{l|c|c}
      \toprule
      \textbf{Method} & PSNR & SSIM \\
      \midrule
      DCP \cite{5567108} & 11.03 & 0.728 \\
      STMRF \cite{Cai2016RealTimeVD} & 15.54 & 0.693 \\
      FDVD \cite{CVPR20_Tassano} & 16.37 & 0.656 \\
      GDN \cite{9010659} & 19.69 & 0.854 \\
      EVDNET \cite{li2017end} & 17.41 & 0.808 \\
      MSBDN \cite{9156921} & 22.01 & 0.876 \\
      FFA \cite{Qin_Wang_Bai_Xie_Jia_2020} & 16.65 & 0.813 \\
      VDN \cite{8492451} & 16.64 & 0.813 \\
      RDNet \cite{zhao2021refinednet} & 16.93 & 0.804 \\
      DAID \cite{shao2020domain} & 19.20 & 0.821 \\
      EDVR \cite{9025464} & 21.22 & 0.874 \\
      PDVD \cite{CoRR21_Li} & 22.69 & 0.875 \\
      CG-IDN \cite{REVIDE} & 23.21 & 0.884 \\
      LRNET \cite{galshetwar2023lrnet} & 23.89 & 0.896 \\
      DSTM \cite{patil2022dual} & 25.53 & 0.894 \\
      DRFNET \cite{galshetwar2024drfnet} & 25.74 & 0.898 \\
      \textbf{CRFNet \cite{galshetwar2024cross}} & \textbf{25.79} & \textbf{0.899} \\
      \bottomrule
    \end{tabular}
    \label{TAB:ref4}
  \end{minipage}

  &

  
  \begin{minipage}{0.4\textwidth}
    \centering
    \captionof{table}{Single task evaluation: Dehazing comparison on DAVIS-2016 and NYU Depth.}
    \begin{tabular}{l|c|c|c|c}
      \toprule
      \textbf{Method} & \multicolumn{2}{c|}{DAVIS-2016} & \multicolumn{2}{c}{NYU Depth} \\
      & PSNR & SSIM & PSNR & SSIM \\
      \midrule
      TCN \cite{9316931} & 16.61 & 0.619 & 18.83 & 0.614 \\
      FFA \cite{Qin_Wang_Bai_Xie_Jia_2020} & 14.19 & 0.650 & 17.74 & 0.715 \\
      MSBDN \cite{9156921} & 15.41 & 0.706 & 16.67 & 0.658 \\
      GCANet \cite{chen2019gated} & 20.31 & 0.728 & 16.93 & 0.650 \\
      RRO \cite{8734728} & 15.09 & 0.760 & 19.47 & 0.842 \\
      FMENet \cite{9198924} & 16.16 & 0.830 & 19.81 & 0.843 \\
      CANCB \cite{9134933} & 16.44 & 0.834 & 20.87 & 0.890 \\
      RDNet \cite{zhao2021refinednet} & 19.38 & 0.788 & 14.85 & 0.561 \\
      DAID \cite{shao2020domain} & 16.71 & 0.776 & 22.63 & 0.876 \\
      DSTM \cite{patil2022dual} & 21.71 & 0.877 & 23.26 & 0.865 \\
      DRFNET \cite{galshetwar2024drfnet} & 22.62 & 0.879 & 23.64 & 0.874 \\
      LRNET \cite{galshetwar2023lrnet} & 22.04 & 0.835 & \textbf{24.87} & \textbf{0.919} \\
      \textbf{CRFNet \cite{galshetwar2024cross}} & \textbf{22.67} & \textbf{0.880} & 23.81 & 0.897 \\
      \bottomrule
    \end{tabular}
    \label{TAB:ref5}
  \end{minipage}

  \end{tabular}}
\end{table*}

We have evaluated and compared current SOTA approaches in terms of quantitative and qualitative results. PSNR and SSIM estimates have been employed to reference-based evaluation analysis. While, NIQE, Entropy, BRISQUE and PIQE are used for non-reference evaluation analysis. Detailed description of the reference and no-reference evaluation metrics are summarized in Table~\ref{TAB: parameter_table}. Temporal coherence plays a crucial role in maintaining the stability of successive video frames and preventing flickering from one frame to the next. To evaluate this, simple measures should be considered to check how similar consecutive frames are. Specifically, the metrics such as temporal warping error (TWE), LPIPS, and inter-frame consistency measures assess stability across successive frames. Architecturally, some standard solutions include recurrent or transformer-based temporal fusion, optical-flow guided alignment modules, memory-augmented networks, and temporal consistency losses that penalize flickering and frame-wise discrepancies.
\subsection{Quantitative Analysis}
    The non-reference evaluation analysis for de-hazing, and de-raining on RTTS \cite{8954466}, and RID \cite{8953352} datasets is provided in TABLE \ref{TAB:NonRef} respectively. The Average NIQE, Entropy, BRISQUE and PIQE are considered as non-reference parameters. 
    Furthermore, the reference based parameter analysis in terms of average PSNR and SSIM is presented in Tables~\ref{TAB:ref3} to~\ref{tab:allinone} across benchmark datasets for various restoration tasks. Three types of methods or models are compared: \textbf{single task methods} that are specialized for a specific degradation such as haze, rain, or snow; \textbf{multi task and multi weather models} that are task-wise fine tuned and evaluated across multiple tasks; and \textbf{all in one} restoration models that are trained once on a combined dataset to handle diverse degradations including blur, noise, low light, and adverse weather within a unified framework. In Table~\ref{TAB:ref3} to Table~\ref{TAB:Ref1}, the first, second, and third partitions along rows correspond to single-task, multi-task/multi-weather, and all-in-one approaches respectively.
    
    \noindent \textbf{Dehazing on Image and Video Benchmarks (Tables~\ref{TAB:ref6}, ~\ref{TAB:ref4}, and ~\ref{TAB:ref5}):}
         results reveal a clear shift from prior-based and CNN-driven dehazing methods toward context-aware and generative architectures. On the Dense Haze dataset, traditional priors~\cite{5567108, wu2023ridcp} perform poorly, while CNN-based methods~\cite{li2017aod, Qin_Wang_Bai_Xie_Jia_2020, wu2021contrastive} show only limited gains. Transformer-based models~\cite{song2023vision, qiu2023mb, wang2022uformer, zamir2022restormer} further improve performance by leveraging global context, with diffusion-based approaches achieving the best results, led by Defusion \cite{Luo_2025_CVPR}. Similar trends are observed on the REVIDE benchmark, where temporally aware models~\cite{patil2022dual, galshetwar2024drfnet, galshetwar2024cross} outperform frame-based methods. On DAVIS 2016 and NYU Depth, depth and motion aware models~\cite{shao2020domain, galshetwar2023lrnet, galshetwar2024cross}, achieve the best balance between PSNR and SSIM. 
        
    \noindent \textbf{Raindrop and Rain Removal (Tables~\ref{TAB:ref3} and ~\ref{TAB:ref7}):} summarize quantitative results for raindrop removal on AGAN Data and rain streak removal on the SPAD dataset.
        Early CNN based methods~\cite{eigen2013restoring, Qu_2019_CVPR, quan2021removing, quan2019deep, xiao2022image} provide steady improvements but remain limited in handling complex rain structures.
        Recent transformer based architectures~\cite{wang2022uformer, zamir2022restormer, zamir2021multi, purohit2021spatially, 10378504, Chen_2023_CVPR, tu2022maxim}, consistently achieve higher PSNR and SSIM by leveraging global context and long-range dependencies.
        Diffusion-based approaches~\cite{qian2018attentive,  ozdenizci2023restoring, zhou2024adapt} further improve restoration quality, with AST-B achieving the best performance on both datasets.
        Overall, the results demonstrate a clear progression from convolutional designs to attention driven and generative models, highlighting the effectiveness of global context modeling for rain related restoration tasks.

    \noindent \textbf{Snow Removal (Table~\ref{TAB:ref8}):} compares snow removal performance across Snow100K, SRRS, and CSD datasets. Early CNN based methods~\cite{zheng2013single, liu2018desnownet} show limited robustness under dense snow. Attention and invertible models~\cite{quan2019deep, chen2020jstasr, quan2023image} significantly improve structural fidelity. Transformer based~\cite{valanarasu2022transweather, Luo_2025_CVPR} further enhance performance on Snow100K. Overall, CCN \cite{cheng2023context} achieves the best and most consistent results across all datasets, demonstrating strong generalization to diverse snow patterns. Also, the results indicate that multi weather generalization benefits significantly from global context modeling and generative priors.

    \noindent \textbf{Multi-task Analysis (Table~\ref{TAB:Ref1}):} shows that DTMIR achieves the best overall performance, particularly in SSIM, with consistent results across haze, snow, and rain. Transformer-based methods are competitive but slightly less robust, highlighting the importance of multi-task-aware parameter tuning.

    \noindent \textbf{All-in-one Restoration (Table~\ref{tab:allinone}):} demonstrate that jointly trained models generalize better across heterogeneous degradations than task-specific baselines. PromptIR achieves a clear gain over AirNet, validating the effectiveness of prompt-guided adaptation in multi-task restoration. DFIR attains the best overall PSNR/SSIM, indicating stronger cross-task consistency. The improvements are particularly notable in deraining, highlighting robustness to complex mixed degradations.

\vspace{-4mm}
\subsection{Qualitative Analysis}
\begin{figure}[t]
\centering
  \includegraphics[width=1\linewidth, keepaspectratio]{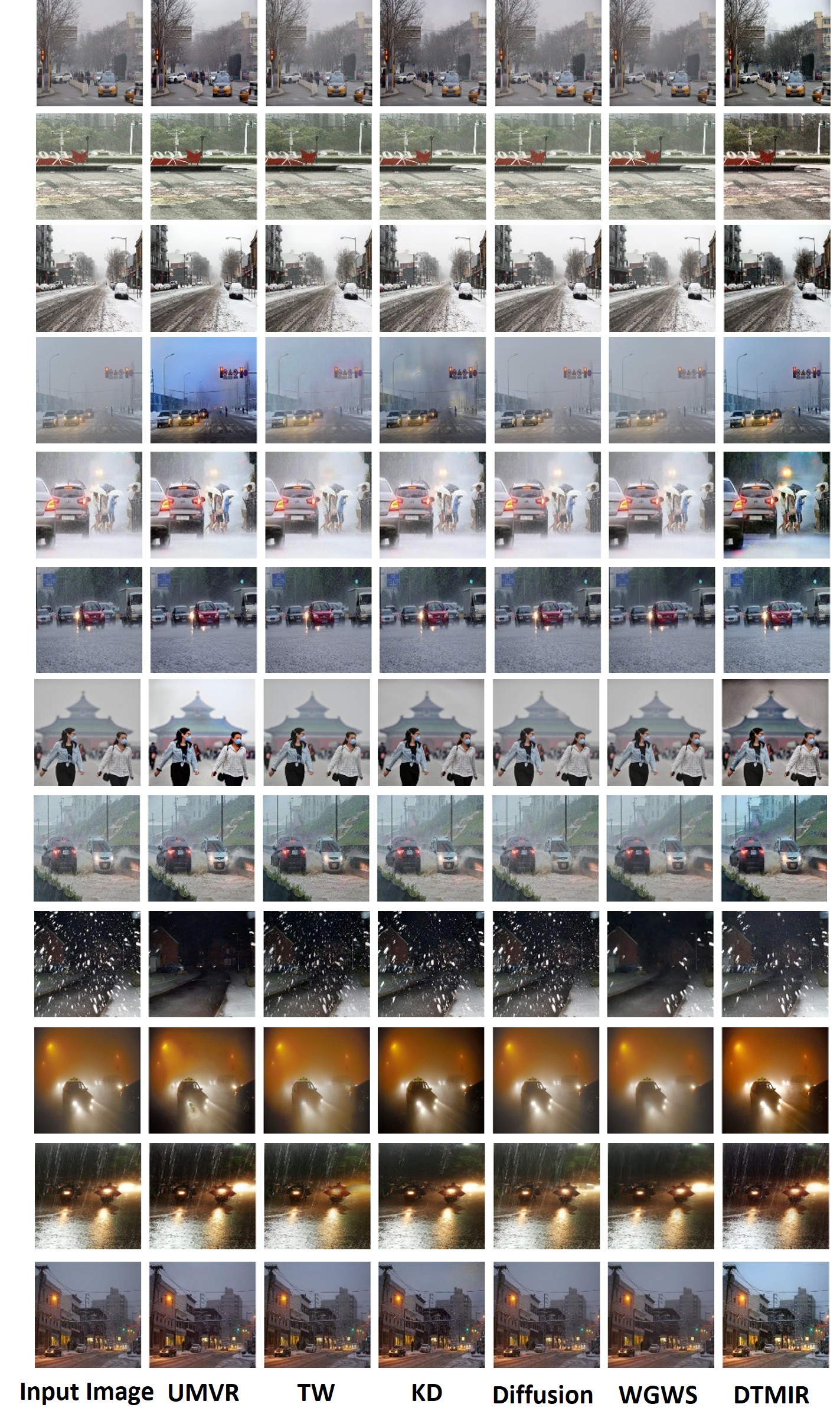}
  \vspace{-7mm}
   \caption{Visual result analysis of the existing methods: UMVR \cite{kulkarni2022unified}, KD \cite{9879902}, TW \cite{valanarasu2022transweather}, Diffusion \cite{ozdenizci2023restoring}, WGWS \cite{10204275} and DTMIR \cite{patil2023multi} on real-world weather degraded images.}
\label{FIG:Qual1}
\end{figure}

In this section, the visual result analysis on day and night-time degraded images is provided. Refer Figure \ref{FIG:Qual1} for day-time and night-time analysis. The UMVR \cite{kulkarni2022unified}, KD \cite{9879902}, TW \cite{valanarasu2022transweather}, Diffusion \cite{ozdenizci2023restoring}, WGWS \cite{10204275} and DTMIR \cite{patil2023multi} methods are considered for visual result analysis purpose. Result analysis for day-time degradations:
\begin{itemize}
    \item \textbf{De-hazing analysis}: Comparison among existing methods is shown in Fig \ref{FIG:Qual1}. We observe the limitations like a lack of sharpness in distant details, artifacts near edges, and reduced visual consistency, color distortions, tends to smooth textures, and over-enhance edges.

    \item \textbf{De-raining analysis}: Figure \ref{FIG:Qual1} shows the visual comparison between the existing methods for image de-raining. However, these methods have limitations, like background smoothing, some streaks and artifacts remain in regions with high rain density, losing texture details, and color tones appear slightly unnatural.

    \item \textbf{De-snowing analysis}: Visual comparison among existing methods shown in Figure \ref{FIG:Qual1} achieved significant performance for snow removal task. However, limitations like over-smooth regions, unnatural smoothing in darker regions, over-saturation, larger snow regions are still need to handle effectively.
\end{itemize}

\begin{table}[t]
  \centering
  \resizebox{\linewidth}{!}{
  \begin{tabular}{cc}

  \begin{minipage}{0.3\textwidth}
    \centering
    \captionof{table}{Quantitative analysis of raindrop removal on AGAN-Data \cite{qian2018attentive}}
    \begin{tabular}{l|c|c}
      \toprule
      \textbf{Method} & PSNR & SSIM \\
      \midrule
      Eigen’s \cite{eigen2013restoring} & 21.31 & 0.757 \\
      Pix2pix \cite{Qu_2019_CVPR} & 27.20 & 0.836 \\
      CCN \cite{quan2021removing} & 31.34 & 0.929 \\
      Quan’s \cite{quan2019deep} & 31.37 & 0.918 \\
      AttenGAN \cite{qian2018attentive} & 31.59 & 0.917 \\
      IDT \cite{xiao2022image} & 31.87 & 0.931 \\
      \midrule
      Uformer \cite{wang2022uformer} & 29.42 & 0.906 \\
      TKLMR \cite{chen2022learning} & 30.99 & 0.927 \\
      DuRN \cite{liu2019dual} & 31.24 & 0.926 \\
      MAXIM-2S \cite{tu2022maxim} & 31.87 & 0.935 \\
      \midrule
      All-in-One \cite{li2020all} & 31.12 & 0.927 \\
      Diffusion128 \cite{ozdenizci2023restoring} & 29.66 & 0.923 \\
      TransWeather \cite{valanarasu2022transweather} & 30.17 & 0.916 \\
      Diffusion64 \cite{ozdenizci2023restoring} & 30.71 & 0.931 \\
      AWRCP \cite{ye2023adverse} & 31.93 & 0.931 \\
      \textbf{AST-B \cite{zhou2024adapt}} & \textbf{32.45} & \textbf{0.937} \\
      \bottomrule
    \end{tabular}
    \label{TAB:ref3}
  \end{minipage}

  &

  
  \begin{minipage}{0.25\textwidth}
    \centering
    \caption{Results of existing methods on SPAD \cite{wang2019spatial} dataset for rain streak removal.}
    \resizebox{\linewidth}{!}{
    \begin{tabular}{l|c|c}
    \toprule
    \multicolumn{1}{c|}{\multirow{2}{*}{Method}} & \multicolumn{2}{c}{SPAD}\\
  
         & PSNR & SSIM\\
    \midrule
    DDN \cite{8099669} & 36.16 & 0.9463 \\
    RESCAN \cite{li2018recurrent} & 38.11 & 0.9797 \\
    PReNet \cite{8953349} & 40.16 &  0.9816 \\
    RCDNet \cite{9157358} & 43.36 &  0.9831 \\
    SPDNet \cite{9710307} & 43.55 & 0.9875 \\
    DualGCN \cite{Fu_article} & 44.18 & 0.9902 \\
    SEIDNet \cite{NEURIPS2022_1dc2fe8d} & 44.96 & 0.9911\\
    Fu et al. \cite{10035447} & 45.03 & 0.9907 \\
    SCD-Former \cite{10378504} & 46.89 & 0.9941 \\
    IDT \cite{xiao2022image} & 47.34 & 0.9929 \\
    \midrule
    SPAIR \cite{purohit2021spatially} & 44.10 & 0.9872 \\
    Restormer \cite{zamir2022restormer} & 46.25 & 0.9911 \\
    MPRNet \cite{zamir2021multi} & 45.00 &  0.9897 \\
    Uformer \cite{wang2022uformer} & 47.84 &  0.9925 \\
    DRSformer \cite{Chen_2023_CVPR} & 48.53 & 0.9924 \\
    \midrule
    \textbf{AST-B \cite{zhou2024adapt}} & \textbf{49.72} & \textbf{0.9944} \\
    \bottomrule
    \end{tabular}}
    \label{TAB:ref7}
    \end{minipage}
     \end{tabular}}
\end{table}

\begin{table}[t]
    \centering
    \caption{Comparative quantitative result analysis of SOTA approaches for snow removal.}
    \resizebox{\linewidth}{!}{
    \begin{tabular}{l|c|c|c|c|c|c}
    \toprule
    \multicolumn{1}{c|}{\multirow{2}{*}{Method}} & \multicolumn{2}{c}{Snow 100K~\cite{liu2018desnownet}} & \multicolumn{2}{c}{SRRS~\cite{Chen2020}} & \multicolumn{2}{c}{CSD~\cite{chen2021all}}\\
  
      & PSNR & SSIM & PSNR & SSIM & PSNR &  SSIM\\
    \midrule
    MGF \cite{zheng2013single} & 22.41 & 0.77 & 15.78 & 0.74 & 13.98 & 0.67 \\
    DesnowNet \cite{liu2018desnownet} & 30.11 & 0.93 & 20.38 & 0.84 & 20.13 & 0.81 \\
    S-Attention \cite{quan2019deep} & 29.94 & 0.89 & 26.56 & 0.90 & 27.85 & 0.88 \\
    JSTASR \cite{chen2020jstasr} & 28.59 & 0.86 & 25.82 & 0.89 & 27.96 & 0.88 \\
    DesnowGAN \cite{jaw2020desnowgan} & 31.11 & 0.95 & - & - & 27.09 & 0.88 \\
    InvDN \cite{liu2021invertible} & 27.99 & 0.81 & 26.49 & 0.88 & 27.46 & 0.86 \\
    HDCW-Net \cite{chen2021all} & 24.10 & 0.80 & 27.78 & 0.92 & 29.06 & 0.91 \\
    InvDSNet \cite{quan2023image} & 32.41 & 0.93 & 29.25 & 0.95 & 31.85 & 0.96 \\
    \textbf{CCN \cite{cheng2023context}} & \textbf{33.64} & \textbf{0.95} & \textbf{37.15} & \textbf{0.99} & \textbf{32.70} & \textbf{0.98} \\
    \midrule
    ResFlow \cite{Qin_2025_CVPR} & 31.86 & 0.917 & - & - & - & - \\
    \midrule
    TransWeathe \cite{valanarasu2022transweather} & 32.06 & 0.94 & 29.05 & 0.95 & 31.13 & 0.95 \\
    Defusion \cite{Luo_2025_CVPR} & 32.11 & 0.926 & - & - & - & - \\
    \bottomrule
    \end{tabular}}
    \label{TAB:ref8}
\end{table}

\noindent The methods UMVR \cite{kulkarni2022unified}, KD \cite{9879902}, TW \cite{valanarasu2022transweather}, Diffusion \cite{ozdenizci2023restoring}, WGWS \cite{10204275} and DTMIR \cite{patil2023multi} are trained on day-time hazy, rainy and snowy degradations. These models are directly tested on night-time degraded images and results are provided in Figure \ref{FIG:Qual1}. From the results, it is clear that the existing SOTA methods are able to handle night-time degradations to some extent. Detailed analysis for night-time degradations is:
\begin{itemize}
    \item \textbf{De-hazing analysis}: The results (refer row 4 and 10 from Figure \ref{FIG:Qual1}) are reasonable for night-time haze removal task. There are various limitations like haze remains in distant areas, brighter regions reduces naturalness, uneven haze removal, introducing minor artifacts in darker regions need to handle effectively.

    \item \textbf{De-raining analysis}: For night-time rain-removal task, the existing methods achieved significant performance (row 5 and 11 from Figure \ref{FIG:Qual1}). However, these methods are suffering from different artifacts, loosing finer details, over-enhancement, handling heavy rain, etc.  

    \item \textbf{De-snowing analysis}: The provided results (row 9 and 12 from Figure \ref{FIG:Qual1}) shows satisfactory performance for night-time snow removal task. But, issues such as over-smoothing, excessive blurring, and reduced texture clarity should be effectively handled. 
\end{itemize}

\begin{table}[t]
\begin{center}
\caption{Reference Parameter Analysis for De-hazing with Rain+haze, De-raining with RainDrop and Snow with SNOW 100K Datasets in terms of Average PSNR/SSIM.}
\resizebox{\linewidth}{!}{
\begin{tabular}{l|llllll}
\toprule
  \multicolumn{1}{c|}{\multirow{2}{*}{Method}} & \multicolumn{2}{c}{RTTS~\cite{8953352}} & \multicolumn{2}{c}{AGAN Data~\cite{qian2018attentive}} & \multicolumn{2}{c}{SNOW 100K~\cite{liu2018desnownet}}\\
  
     & PSNR & SSIM & PSNR & SSIM & PSNR & SSIM\\
\midrule
AOD-Net \cite{li2017aod} (ICCV-17) & 24.71 & 0.898 & 31.12 & 0.927 & 28.33 & 0.882 \\
\midrule
Restormer \cite{zamir2022restormer} (CVPR-22) & 27.24 & 0.920 & 29.29 & 0.937 & 27.76 & 0.906 \\
MPRNet \cite{zamir2021multi} (CVPR-21) & 28.08 & 0.931 & 29.45 & 0.941 & 27.92 & 0.911 \\
Uformer \cite{wang2022uformer} (CVPR-23) & 25.40 & 0.889 & 27.38 & 0.919 & 26.60 & 0.887 \\
LPM \cite{10453462} (TIP-24) & 28.68 & 0.940 & 30.40 & 0.956 & 28.54 & 0.922 \\
ResFlow \cite{Qin_2025_CVPR} (CVPR-25) & - & - & 32.82 & 0.936 & 31.86 & 0.917 \\
\midrule
TW \cite{valanarasu2022transweather} (CVPR-22) & 28.83 & 0.900 & 30.17 & 0.916 & 29.31 & 0.888 \\
KD \cite{9879902} (CVPR-22) & 24.20 & 0.904 & 30.47 & 0.954 & 26.96 & 0.897 \\
WeaFU \cite{article} (TCSVT-24) & - & - & - & - & 29.49 & 0.920 \\
MWFormer \cite{10767188} (TIP-24) & 30.27 & 0.912 & 31.91 & 0.927 & 30.92 & 0.909 \\
Diffusion \cite{ozdenizci2023restoring} (TPAMI-23) & 29.64 & 0.931 & 30.71 & \textbf{0.931} & 30.09 &  0.904 \\
MWCNet \cite{10697214} (TCSVT-25) & \textbf{30.78} & \textbf{0.949} & 31.18 & 0.940 &  30.92 & 0.923 \\
Defusion \cite{Luo_2025_CVPR} (CVPR-25) & - & - & \textbf{33.81} & \textbf{0.967} & \textbf{32.11} & \textbf{0.926} \\

\bottomrule
\end{tabular}}
\label{TAB:Ref2}
\end{center}
\vspace{-4mm}
\end{table}

\begin{table}[t]
\begin{center}
\caption{Reference Parameter Analysis for De-hazing with SOTS, De-raining with ORD and Snow with CSD Datasets in terms of Average PSNR/SSIM.}
\setlength{\tabcolsep}{3.5pt}
\resizebox{\linewidth}{!}{
\begin{tabular}{l|llllll}
\toprule
  \multicolumn{1}{c|}{\multirow{2}{*}{Method}} & \multicolumn{2}{c}{SOTS~\cite{li2019benchmarking}} & \multicolumn{2}{c}{CSD~\cite{chen2021all}} & \multicolumn{2}{c}{ORD~\cite{6353522}}\\
  
     & PSNR & SSIM & PSNR & SSIM & PSNR & SSIM\\
\midrule
UMVR \cite{kulkarni2022unified} (TMM-22) & 33.41 & 0.980 & 28.65 & 0.900 & 22.99 & 0.830 \\
DTMIR \cite{patil2023multi} (ICCV-23) & \textbf{36.26} & \textbf{0.987} & \textbf{32.95} & \textbf{0.942} & \textbf{31.24} & \textbf{0.951} \\
\midrule
KD \cite{9879902} (CVPR-22) & 34.64 & 0.985 & 31.35 & 0.950 & 29.05 & 0.916 \\
TransWeather \cite{valanarasu2022transweather} (CVPR-22) & 32.45 & 0.955 & 29.76 & 0.940 & 27.96 & 0.950 \\
\bottomrule
\end{tabular}}
\label{TAB:Ref1}
\end{center}
\vspace{-4mm}
\end{table}

\begin{table}[ht]
\centering
\caption{Comparisons under All-in-one restoration setting: single model trained on a combined set of images originating from different degradation types.}
\label{tab:allinone}
\resizebox{\linewidth}{!}{
\begin{tabular}{l|cccc}
\toprule
\textbf{Method} & \textbf{Dehazing} & \textbf{Deraining} & \multicolumn{2}{c}{\textbf{Denoising on BSD68 dataset}} \\
 & SOTS~\cite{li2019benchmarking} & Rain100L~\cite{7780668} & $\sigma=25$ & $\sigma=50$ \\
\midrule
BRDNet~\cite{TIAN2020461} (NN-20) & 23.23/0.895 & 27.42/0.895 & 29.76/0.836 & 26.34/0.836 \\
LPNet~\cite{8953950} (CVPR-19) & 20.84/0.828 & 24.88/0.784 & 24.77/0.748 & 21.26/0.552 \\
NAFNet~\cite{10.1007/978-3-031-20071-7_2} (ECCV-22) & 24.11/0.960 & 33.64/0.956 & 30.47/0.865 & 27.12/0.754 \\
FDGAN~\cite{dong2020fd} (CVPR-20) & 24.71/0.924 & 29.89/0.933 & 28.81/0.868 & 26.43/0.776 \\
MPRNet~\cite{zamir2021multi} (CVPR-21) & 25.28/0.954 & 33.57/0.954 & 30.89/0.880 & 27.56/0.779 \\
DL~\cite{8750830} (TPAMI-21) & 26.92/0.391 & 32.62/0.931 & 30.41/0.861 & 26.90/0.740 \\
Restormer~\cite{zamir2022restormer} (CVPR-22) & 27.78/0.958 & 33.78/0.958 & 30.67/0.865 & 27.63/0.792 \\
AirNet~\cite{9879292} (CVPR-22) & 27.94/0.962 & 34.90/0.967 & 31.26/0.888 & 28.00/0.797 \\
Perceive-IR~\cite{10990319} (TIP-25) & 30.87/0.975 & 38.29/0.980 & 31.53/0.890 & 28.31/0.804 \\

NDR~\cite{10680296} (TIP-24) & 28.64/0.962 & 35.42/0.969 & 31.36/0.887 & 28.10/0.798 \\

IDR~\cite{10204072} (CVPR-23) & 29.87/0.970 & 36.03/0.971 & 31.32/0.884 & 28.04/0.798 \\

Gridformer~\cite{wang2024gridformer} (IJCV-24) & 30.37/0.970 & 37.15/0.972 & 31.37/0.887 & 28.11/0.801 \\

VLU-Net~\cite{11094978} (CVPR-25) & 30.71/0.980 & 38.93/0.984 & 31.48/0.892 & 28.23/0.804 \\
PromptIR~\cite{potlapalli2023promptir} (ANIPS-23) & 30.58/0.974 & 36.37/0.972 & 31.31/0.888 & 28.06/0.799 \\
AdaIR~\cite{cui2024adair} (arXiv-24) & 31.06/0.980 & 38.64/0.983 & 31.45/0.892 & 28.19/0.802 \\
MoCE-IR~\cite{Zamfir_2025_CVPR} (CVPR-25) &  31.34/0.979 & 38.57/\textbf{0.984} & 31.45/0.888 & 28.18/0.800 \\
InstructIR~\cite{conde2024high} (ECCV-24) & 30.22/0.959 & 37.98/0.978 & 31.52/0.890 & 28.30/0.804 \\
Cat-AIR~\cite{jiang2025cataircontenttaskawareallinone} (arxiv-25) & 31.49/0.980 & 38.43/0.983 & 31.44/0.892 & 28.14/0.803 \\
DFPIR~\cite{Tian_2025_CVPR} (CVPR-25) & 31.87/0.980 & 38.65/0.982 & 31.47/0.893 & 28.25/0.806 \\
SE-SymUNet~\cite{jiao2025unleashingdegradationcarryingfeaturessymmetric} (arxiv-25) & \textbf{32.02/0.983} & \textbf{39.23/0.986} & \textbf{31.58/0.895} & \textbf{28.33/0.809} \\

\bottomrule
\end{tabular}}
\end{table}
\begin{table}[t]
\begin{center}
\caption{Computational Complexity analysis of SOTA Methods in Terms of Number of Trainable Parameters, FLOPs and Inference Time (sec).}
\resizebox{\linewidth}{!}{
\begin{tabular}{l|lllcc}
\toprule
Methods & Parameters & Resolution & GFLOPS & Inference Time (msec)\\
\midrule
LPM \cite{10453462} (TIP-24) & 126M & 480$\times$640 & 51.5 & 0.09s \\

Diffusion \cite{ozdenizci2023restoring} (TPAMI-23) & 110M  & 640$\times$432 & 475.43 & 20.52s\\
Uformer \cite{wang2022uformer} (CVPR-23) & 51M  & 512$\times$512 & 357.8 & 0.33s\\
MPRNet \cite{zamir2021multi} (CVPR-21) & 16M & 480$\times$640 & 6534 & 0.53s \\
TransWeather \cite{valanarasu2022transweather} (CVPR-22) & 38M & 256$\times$256 & 38.1 & 0.14s\\
MoCE-IR\cite{Zamfir_2025_CVPR} (CVPR-25) & 25.35M  & 256$\times$256 & 88.42 & 0.220s \\
DTMIR \cite{patil2023multi} (ICCV-23) & 29.7M & 256$\times$256 & 279.54 & 0.222s\\
AirNet \cite{9879292} (CVPR-22) & 8.93M & 256$\times$256 & 238 & 0.451s \\
MSBDN \cite{inproceedings2} (CVPR-22) & 28.71M  & 256$\times$256 & 24.56 & 0.059s \\
WGWS \cite{10204275} (CVPR-23) & 5.97M & 256$\times$256 & 1.68 & 0.03s\\
ACL \cite{Gu_2025_CVPR} (CVPR-25) & 4.6M & 256$\times$256 & 55 & 0.2s\\

PromptIR \cite{potlapalli2023promptir} (NeuRIPS-23) & 35.59M  & 256$\times$256 & 158.20 & 0.291s \\

\bottomrule
\end{tabular}}
\label{TAB:para1}
\end{center}
\end{table}


\subsection{Computational Complexity Analysis}
Any image or video restoration method act as a pre-processing step for high vision tasks like object detection, activity recognition \cite{chaudhary2019depth, chaudhary2019deep}, etc. Therefore, maintaining effective computational complexity is important aspect for real-world applications. The TABLE \ref{TAB:para1} shows the computational complexity analysis of the existing multi-weather methods in terms of number of trainable parameters, size and GFLOPs. The number of trainable parameters, size of the DTMIR \cite{patil2023multi} is 11M and 44.01MB respectively which is less than all other methods. The GFLOPs of TW \cite{valanarasu2022transweather} is 12.24G which is less than all other methods.

\noindent \textbf{Real-Time Readiness Analysis:} 
The results show that model size and FLOPs alone do not reflect real-time readiness; inference latency is the primary constraint. Methods like LPM, MSBDN, WGWS achieve practical real-time readiness with sub-second inference, while diffusion and all-in-one models face prohibitive latency. Overall, efficient architectures and task-aware optimization are critical for real-time deployment.



\section{Research Needs and Future Directions}
\subsection{Research Needs}
\begin{itemize}
    \item \textbf{Comprehensive Benchmark Datasets:} Large-scale, high-resolution datasets covering rain, snow, haze, fog, and dust across intensities and regions are still needed. These should include image and video sequences with temporally consistent annotations for robust evaluation across day/night transportation scenarios.
    
    \item \textbf{Generalized Multi-weather Restoration Models:} While recent methods handle multiple degradations, generalization to unseen or compound conditions remains limited. Improving scalability under dynamic mixed weather requires stronger architectures trained on diverse data.
    
    \item \textbf{Real-time and Edge-Efficient Processing:} Smart transportation requires fast edge inference. Research should focus on quantization, pruning, distillation, etc., to balance speed, efficiency, and accuracy.
    
    \item \textbf{Domain Adaptation and Synthetic-to-Real Generalization:} Models trained on synthetic data often fail in real-world settings. Domain adaptation and self-supervised methods (e.g., unsupervised adaptation, style transfer, GAN-based learning) are needed to bridge this gap.
\end{itemize}

\subsection{Future Directions}
\begin{itemize}

    \item \textbf{Unified All-Weather Restoration and Downstream Perception:} 
    Future research should move beyond treating restoration as an isolated pre-processing step and instead focus on unified frameworks that jointly address all-weather restoration and downstream vision tasks. While early efforts integrate restoration with object detection \cite{10636782, 10012056}, they are limited in object categories and task diversity. Extending such approaches to a wider range of perception tasks including depth estimation \cite{hambarde2020s2dnet}, activity recognition \cite{chaudhary2018tsnet}, and scene understanding, remains largely unexplored. Moreover, developing weather-robust models that directly perform these tasks on degraded images and videos, without explicit restoration, is a promising direction for real-time and safety-critical applications such as ADAS and autonomous driving.

    \item \textbf{Adaptive, Efficient, and Real-time All-Weather Video Restoration:} 
    Existing multi-weather restoration methods are predominantly image-based, with limited attention to video restoration \cite{patil2022dual, galshetwar2022consolidated}. Future work should emphasize temporally consistent, real-time video restoration capable of handling mixed and compound degradations. This requires weather-adaptive and degradation-aware neural architectures that dynamically adjust to unknown environmental conditions using mechanisms such as self-attention, reinforcement learning, or uncertainty-aware modeling. At the same time, practical deployment demands lightweight and energy-efficient designs using model compression techniques such as pruning, weight sharing, tensor decomposition, and low-bit quantization, particularly for embedded and automotive platforms.

    \item \textbf{Develop Benchmarks with Compound Degradations under Dynamic Lighting Conditions:} 
    A critical bottleneck in advancing all-weather vision systems is the lack of standardized benchmark datasets that reflect real-world complexity. Future efforts should focus on creating realistic datasets that model compound degradations, such as simultaneous haze, rain, blur, and noise under dynamic lighting conditions caused by headlights, streetlights, reflections, shadows, and rapidly changing illumination. Such benchmarks will enable rigorous evaluation, support the development of more robust and generalizable models, and better align research progress with real-world autonomous and intelligent transportation scenarios.

\end{itemize}
\section{Conclusion}
The significant challenges posed due to adverse weather conditions for various applications like autonomous driving, surveillance, and remote sensing are discussed. Further, the development of weather specific and multi-weather restoration approaches with specific limitations from traditional to learning (CNNs, GANs, Transformer, Diffusion, Knowledge distillation and Multimodal) based techniques are discussed. Recent advancements with learning techniques have demonstrated superior performance by capturing complex features. However, challenges related to more diverse datasets, limited video-based multi-weather restoration, all weather object detection with diverse categories still need to be addressed. This survey aims to guide future research efforts, encouraging innovation in multi-weather video restoration and all-weather object recognition to enhance visibility and safety across diverse domains.

\bibliographystyle{IEEEtran}
\bibliography{egbib}

\begin{IEEEbiography}
[{\includegraphics[width=1in,height=2in,clip,keepaspectratio]{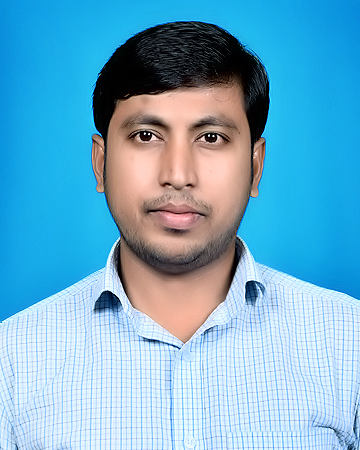}}]
{Vijay M. Galshetwar} received Ph.D. from Punjab Engineering College, Chandigarh, India in 2024. He is currently an Assistant Professor at Finolex Academy of Management and Technology, Ratnagiri, India. His research interests include computer vision, deep learning, and multi-weather image restoration.
\end{IEEEbiography}
\vskip 15pt plus -5fil
\begin{IEEEbiography}
[{\includegraphics[width=1in,height=4in,clip,keepaspectratio]{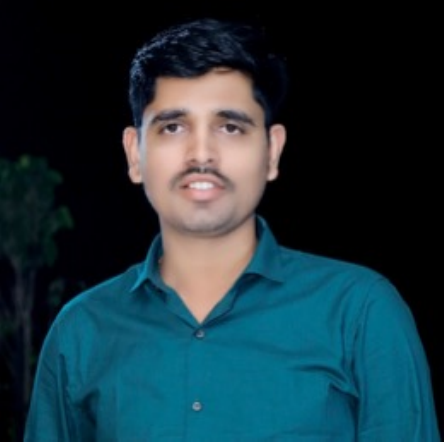}}]
{Praful Hambarde} (Member, IEEE) received Ph.D. from Indian Institute of Technology Ropar in 2022. He is currently an Assistant Professor at the Centre for Artificial Intelligence and Robotics, Indian Institute of Technology Mandi, India. His research interests include computer vision, generative AI, deepfake forensics, robotics, and medical image analysis.
\end{IEEEbiography}
\vskip 25pt plus -1fil
\begin{IEEEbiography}
[{\includegraphics[width=1in,height=2in,clip,keepaspectratio]{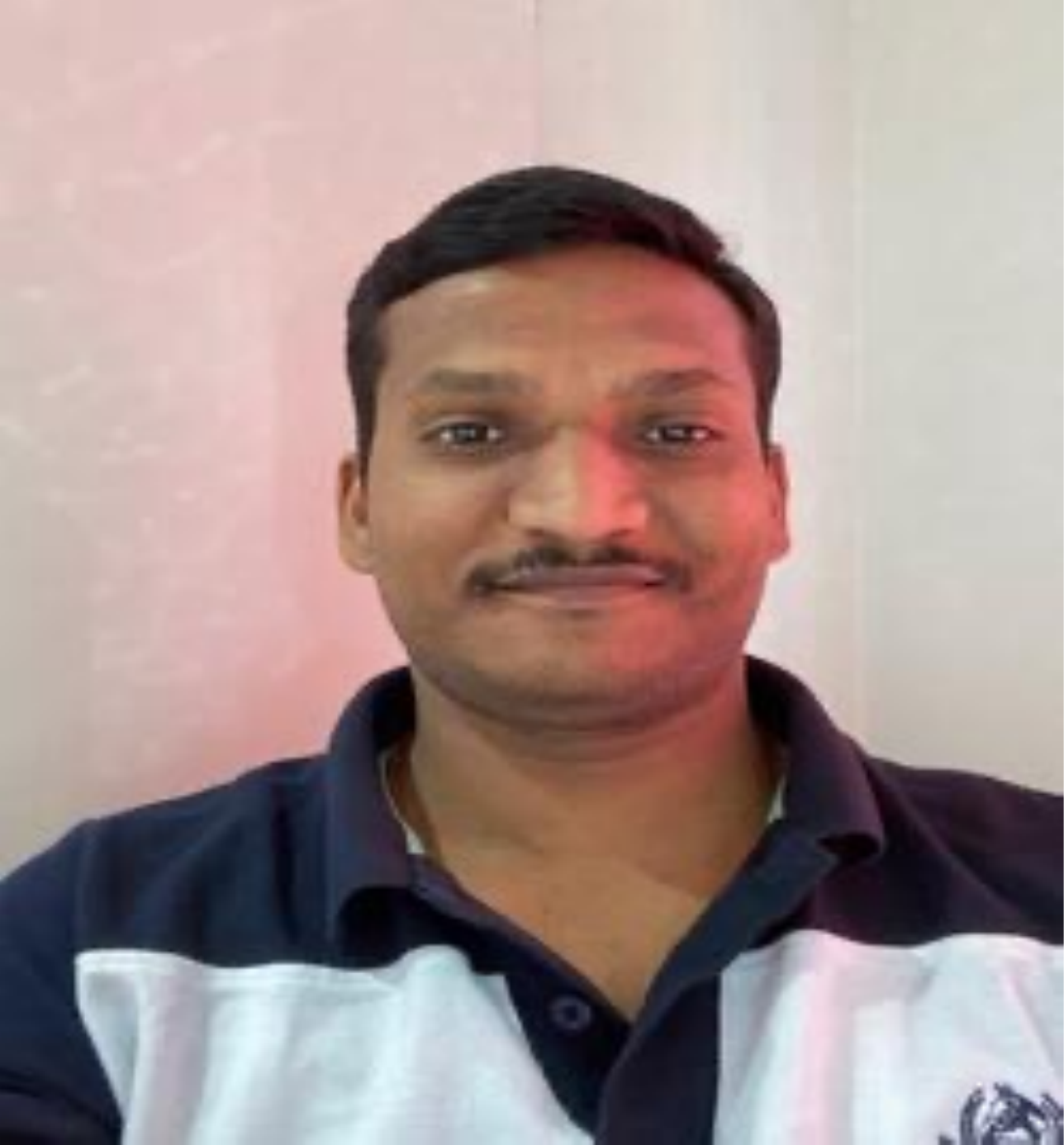}}]
{Prashant W. Patil} received the Ph.D. degree from Indian Institute of Technology Ropar in 2021. He is currently an Assistant Professor with the Mehta Family School of Data Science and Artificial Intelligence, Indian Institute of Technology Guwahati, India. His research interests include Computer Vision, Image/video Restoration, All-weather Object Detection, Multi-modal Deepfake Detection, Underwater Super-resolution,
\end{IEEEbiography}
\vskip 25pt plus -1fil
\begin{IEEEbiography}
[{\includegraphics[width=1in,height=2in,clip,keepaspectratio]{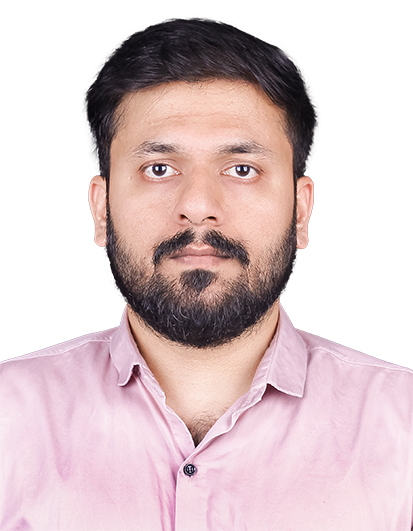}}]
{Akshay Dudhane} received the Ph.D. degree from the Indian Institute of Technology Ropar, India, in 2021. He is currently a Senior Data Scientist at SPACE42, Abu Dhabi, UAE. His research focuses on burst image restoration and enhancement, medical image analysis, satellite image processing, and agentic AI, with an emphasis on building efficient and scalable intelligent systems.
\end{IEEEbiography}
\vskip 25pt plus -1fil
\begin{IEEEbiography}
[{\includegraphics[width=1in,height=2in,clip,keepaspectratio]{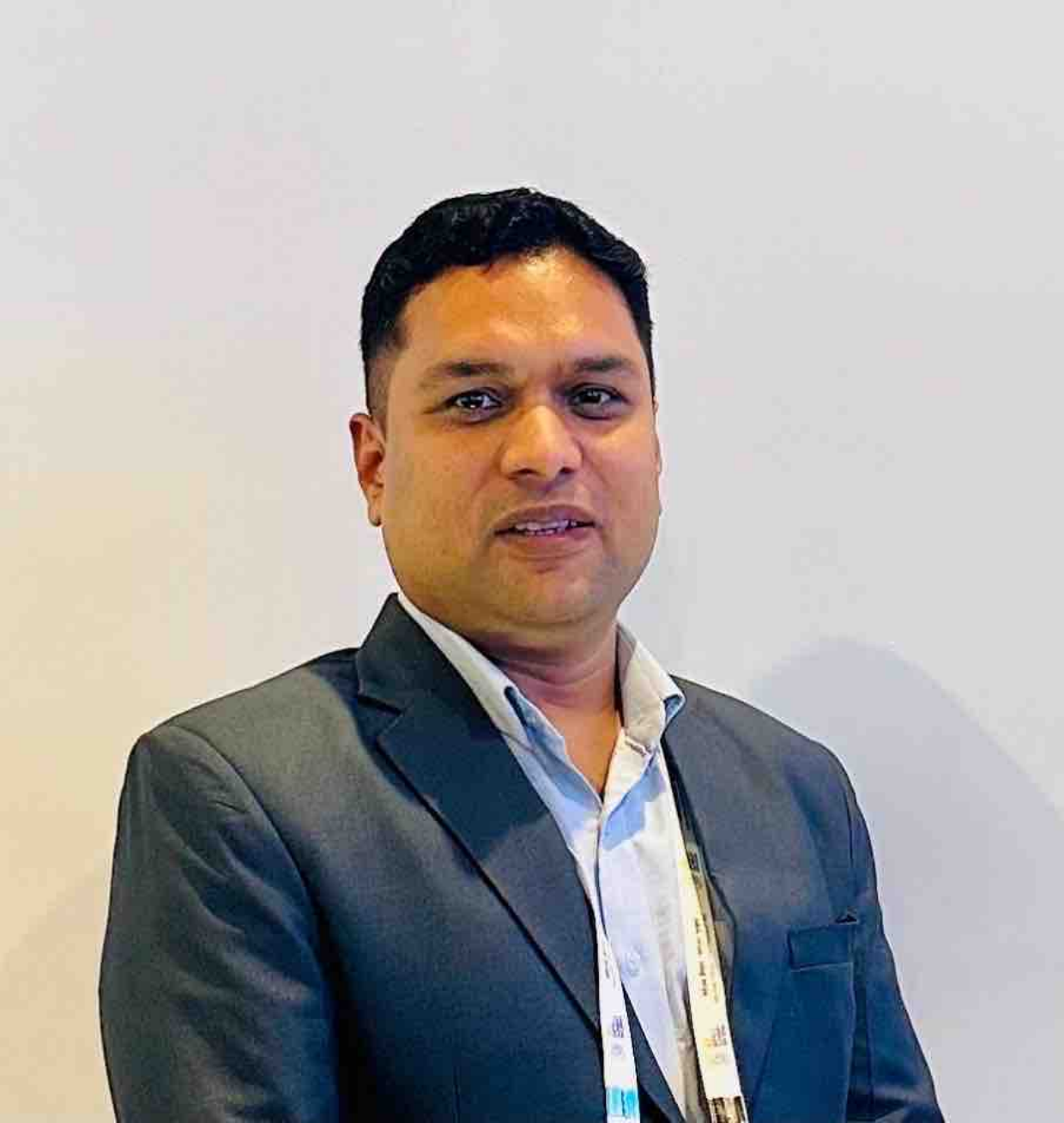}}]
{Sachin Chaudhary} received his Ph.D. from Indian Institute of Technology Ropar, India, in 2019, and was a Post-Doctoral Researcher at Nanyang Technological University (NTU), Singapore. He is currently a Senior Associate Professor with the School of Computer Science, UPES, Dehradun, India. His research interests include intelligent transportation systems, computer vision, and deep learning, focusing on multimodal perception, identity verification, media authenticity, and trustworthy AI.
\end{IEEEbiography}

\end{document}